\documentclass{article}
\usepackage[a4paper, total={7in, 8.5in}]{geometry}

\usepackage{graphicx}
\usepackage{float}
\usepackage{caption}
\usepackage{subcaption}
\usepackage{authblk}
\usepackage{amssymb}
\usepackage{amsmath}

\newtheorem{definition}{Definition}

\usepackage[section]{placeins}
\begin{document}

\title{Effect of Input Noise Dimension in GANs}


\author{Padala Manisha\thanks{manisha.padala@research.iiit.ac.in}}
\author{Debojit Das\thanks{debojit.das@research.iiit.ac.in}}
\author{Sujit Gujar\thanks{sujit.gujar@iiit.ac.in}}
\affil{International Institute of Information Technology, Hyderabad}





\maketitle
\date{}
\begin{abstract}
Generative Adversarial Networks (GANs) are by far the most successful generative models. Learning the transformation which maps a low dimensional input noise to the data distribution forms the foundation for GANs. Although they have been applied in various domains, they are prone to certain challenges like mode collapse and unstable training.  To overcome the challenges, researchers have proposed novel loss functions, architectures, and optimization methods. In our work here, unlike the previous approaches, we focus on the input noise and its role in the generation. 

We aim to quantitatively and qualitatively study the effect of the dimension of the input noise on the performance of GANs. For quantitative measures, typically \emph{Fr\'{e}chet Inception Distance (FID)} and \emph{Inception Score (IS)}  are used as performance measure on image data-sets. We compare the FID and IS values for DCGAN and WGAN-GP. We use three different image data-sets -- each consisting of different levels of complexity. Through our experiments, we show that the right dimension of input noise for optimal results depends on the data-set and architecture used. We also observe that the state of the art performance measures does not provide enough useful insights. Hence we conclude that we need further theoretical analysis for understanding the relationship between the low dimensional distribution and the generated images. We also require better performance measures.
\end{abstract}


\begin{figure}[!htb]
\begin{center}
   \includegraphics[width=\textwidth]{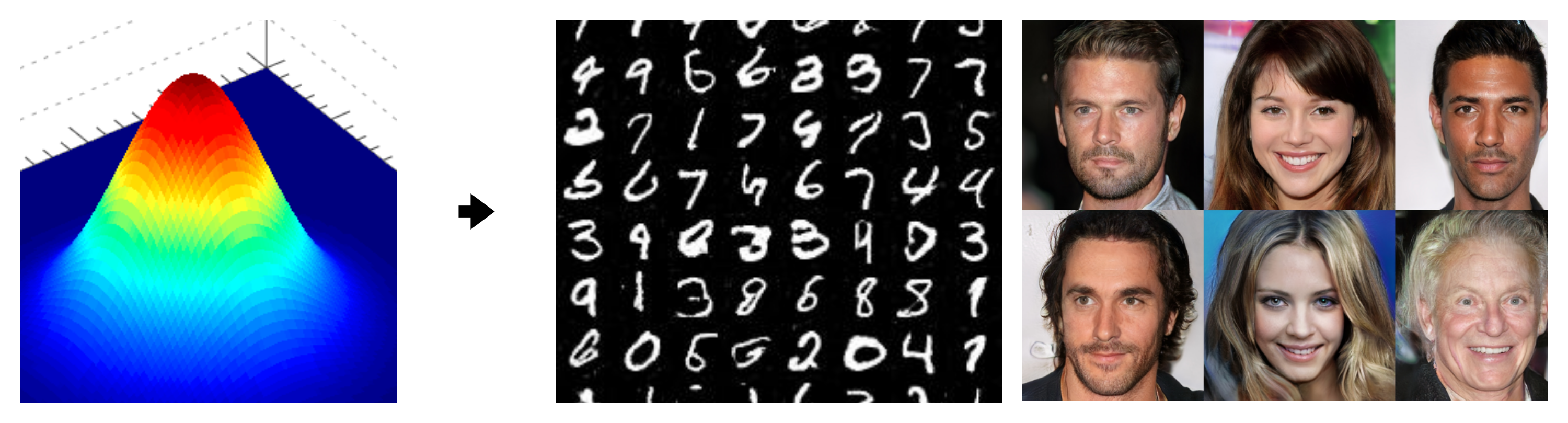}
  \caption{Input Noise to Generated Images \cite{progressive}}
  \label{fig:teaser}
\end{center}{}
\end{figure}



\section{Introduction}
\label{sec:intro}
Generative Adversarial Networks (GANs) are generative models, which learn data distribution and generate samples similar to the real data. These models are useful in learning representations of data without supervision. Deep neural networks are extensively being used for discriminative, retrieval, and clustering tasks. These networks perform well only with a large amount of labeled data. To avoid manually labeling the data, we can transfer the learned representations from the generative models to improve performance for the above tasks. Besides, we can use generative models for simple data augmentation. GANs have enormous applications in a variety of fields. In vision, GANs are used for super-resolution of images \cite{progressive,srgan}, transferring domain knowledge from images of one domain to another \cite{cyclegan, discogan, pix2pix}, object detection \cite{perceptual}, image editing \cite{imageblend}, medical images \cite{anomaly}. In reinforcement learning, GANs are used for generating an artificial environment. One can perform semi-supervised learning using GANs when labels are missing. Other applications include generation of music \cite{musegan}, paintings \cite{paint} and text \cite{seqgan, advnl}. These are but a few of the applications that developed recently.

In a typical GAN set-up, we provide a low dimensional random noise vector as an input, which it maps to a specific data sample having a much higher dimension. Hence given a trained GAN model and simple noise distribution, we can sample from the data distribution without knowing the actual distribution itself.
GANs can be considered as an experimental success. The novelty and simplicity in its set-up contribute to its popularity. GANs lack of a satisfying theoretical explanation has piqued the interest of the research community.  We can consider that GAN learns a lower-dimensional representation of the high dimensional data input and its mapping from lower-dimensional to real-like data sample; whenever we provide a random noise as input.  According to the manifold hypothesis, the real-world high dimensional data lie on low dimensional manifolds embedded in high dimensional space \cite{manifold}. Hence, we believe that the data can have an efficient lower-dimensional representation. Given access to high dimensional distribution of the data, we could efficiently perform dimensionality reduction using existing methods such as PCA \cite{pca}, JL transform \cite{jl}, or deep auto-encoders. However, the main challenge is that we do not know the higher dimensional distribution of the data, and hence mapping becomes difficult. The GAN set-up provides us with an innovative way of mapping the low dimensional noise to the data.

In the vanilla GAN~\cite{ganGoodfellow}, the model primarily has two networks, the \emph{generator} and the \emph{discriminator}. The generator, as its name suggests, is responsible for generating samples that look like real data from a lower-dimensional input noise. The discriminator is used to classify the images generated by the generator and the actual images. While training, the discriminator is trained to improve its ability to discriminate real vs. fake images while the generator is trained to fool the discriminator. The set-up is similar to a two-player zero-sum game, and at equilibrium, the generator samples data points from the data distribution. That is, it learns the data distribution.

Despite the success, GANs face major issues such as mode collapse . As a result of mode collapse, the generator maps multiple noise variables to just one data point. Although the images generated are very sharp and realistic, they do not have large diversity, which we expect in real images. The other issue is that the training is not smooth and may not converge sometimes. There has been quite a lot of research to overcome mode collapse and stabilize the training. Researchers have proposed new losses that guarantee better convergence, new architectures, and new optimizers for overcoming the above challenges. WGAN \cite{arjovskyWGAN}, is the most popular modification over the vanilla GAN which tries to address the challenges faced by changing the loss function. Another major challenge for generative models, in general, is qualitatively estimating their performance. We need to evaluate the quality as well as the diversity of the samples generated. For our analysis we use two measures that are widely used namely, \emph{Fr\'{e}chet Inception Distance (FID)} \cite{ttur} and \emph{Inception Score (IS)} \cite{goodfellow16}. The survey \cite{measures} provides a detailed analysis of the existing measures. 

There have also been papers that try to develop a theoretical framework to analyze the models better. In \cite{manisha}, there are further details about the issues and novel approaches. To best of our knowledge, there is no prior work that discusses the effect of the dimension of the input noise on the GAN performance. Considering that the model aims at mapping the input noise to the data, it is crucial to find the effect of changing the input noise on the performance of the model.

\noindent\emph{Our contribution:}
In this work, we focus on studying the effect of noise on the performance of the model. We vary the dimension of the input noise and its distribution and study its effect on the samples generated for three different data-sets i) Gaussian Data ii) MNIST digits data \cite{mnist} iii) CelebA face data \cite{celeba} (a) 32x32 and (b) 64x64. We use two different kinds of GANs i) DCGAN \cite{dcgan} ii) WGAN-GP \cite{gularajani}. We provide the following results,
 \begin{enumerate}
     \item Quantitative estimation by comparing the FID and IS for the generated samples.
     \item Qualitative estimation: by comparing the samples of images generated after the training converges. 
 \end{enumerate}{}
 We believe that such an analysis would lead not only to the best set of parameters but also to provide useful insights into the working of the model. 
 
 \noindent\emph{Organization:} We discuss the related work in Section \ref{sec:rw}, which is followed by the preliminaries in Section \ref{sec:prelim}, where we discuss the GAN models and performance measures used in detail. Next, we describe the experimental set-up and compare the result in Section \ref{sec:expt}. Finally, we discuss the insights derived from the experiments in Section \ref{sec:disc} before concluding in Section \ref{sec:conc}.
\section{Related Work}
\label{sec:rw}
Restricted Boltzmann Machines (RBMs) \cite{hinton2006} , Deep Belief Networks (DBNs) \cite{dbn}, Variational Autoencoders (VAEs) \cite{kingma2014} are generative models popular before GANs. They have an elegant theory, but they fail to produce complex images when trained on other data-sets such as CIFAR, SVHN, etc. GANs generate images that are sharper and realistic. DCGAN \cite{dcgan} introduced changes in the architecture and hyperparameters that stabilized the training of vanilla GANs on images to a large extent. 
 Vanilla GANs suffer from the issues of \textit{vanishing gradients}, as discussed in \cite{arjovsky01}. In \cite{arjovskyWGAN}, the authors propose Wasserstein GAN (WGAN) to overcome the issue.  In \cite{gularajani}, the authors propose WGAN with gradient penalty (WGAN-GP) for further stability.

Researchers have followed various approaches to resolve the challenges in GANS. Some approaches propose new loss functions\cite{lsgan, ls-gan, mmd}, reforms in architecture \cite{eb-gan, d2gan}, and changes in the optimizer \cite{geiger06, optimism}. There are other papers \cite{ttur, DRAGAN, Arora03} which provide rigorous theoretical analysis. For further details, one can refer to the survey \cite{manisha}, which briefly discusses the reforms introduced by papers to solve the issues discussed. 

In this work, our focus is primarily on the latent input noise, and we study its effect on the generated samples. Few papers modify the input noise and introduce a set of latent variables to control the generated images. The authors in \cite{infogan} made the latent variable represent some visual concepts in an unsupervised manner, and with supervision in \cite{bigbigan, random}. To the best of our knowledge, no paper rigorously explores this aspect. In the next section, we describe the models and performance measures. 
\section{Preliminaries}
\label{sec:prelim}
In this section, we introduce the basic notations required. We also describe the two models and performance measures that we use. The main components of a GAN set-up include,
\begin{enumerate}
    \item[a.] architecture which includes the discriminator $D$ parameterized by $\theta$ and generator $G$ parameterized by $\phi$, 
    \item[b.]  loss function denoted by $V(D_{\theta}, G_{\phi})$,
    \item[c.] optimizer.
\end{enumerate}{}

We assume that the data distribution is $p_d$, and the noise is sampled from another distribution $p_{z}(z)$. The generated samples follow the distribution $p_g$, which is referred to as the model distribution.
In a typical architecture model, we sample $z \sim p_z$ and feed it to the $G$. $z$ is usually low dimensional and follows a simple distribution. When dealing with images, $G$ is a convolutional multi-layered network that takes the $z$ and generates a vector $\hat{x}$, which has the same dimension as the data $x$. $\hat{x} \sim p_g$ is the distribution learnt by $G$. The weights/parameters of $G$ are denoted by $\phi$. The other network is the $D$, parameterized by $\theta$. It takes either $\hat{x}$ or $x$ as input and outputs a single value. The value is a score for the input; a higher score indicates that the image is likely to be sampled from $p_d$ and not $p_g$. The main challenge is to construct a suitable loss and optimize over the loss for achieving our objective of generating realistic images. In this paper, we consider two different kinds of loss i) DCGAN ii) WGAN-GP as further elaborated below,

\subsection{DCGAN}

DCGAN is a modification of vanilla GAN as introduced in \cite{ganGoodfellow}; the authors set the problem as a two-player zero-sum game. $D$ minimizes the classification loss such that it can classify the samples from $p_d$ and $p_g$ differently. While $G$ tries to maximize the same loss or generate samples to fool the $D$. The objective is a simple binary cross-entropy loss used for 2 class classification given below,
\begin{equation}
\label{eq:gan_loss1}
\begin{split}
     \underset{\phi}{min}\ \underset{\theta}{max}
\ V_{G}(D_{\theta},G_{\phi}) =  \  & \mathbb{E}_{x \sim p_{d}(x)}[log D_{\theta}(x)] + \\
 &\mathbb{E}_{z \sim p_{z}(z)}[log(1-D_{\theta}(G_{\phi}(z)))]
\end{split}{}
\end{equation}{}

Samples from $p_d$ are given label 1 and $\hat{x}\sim p_g$ are given 0. The above objective is equivalent to minimizing the \emph{Jenson Shannon Divergence (JSD)} between $p_d$ and $p_g$.

Early in training, when the samples generated by $G$ are very noisy, the discriminator can classify with high confidence, causing the gradients w.r.t. $\phi$ to be very small. Hence, there is no learning signal for $G$ to improve; hence the authors propose to use the following loss for $G$,

\begin{equation}
\label{eq:gan_loss2}
\underset{{\phi}}{max} \ \log(D_{\theta}(G_{\phi}(z)))
\end{equation}
Given the loss, it is challenging to balance the optimization of both $G$ and $D$. The optimal solution lies at the saddle point denoted by $(\phi^*, \theta^*)$. Using \textit{simultaneous gradient descent}, the authors prove the convergence of the loss under specific assumptions. In this method, $\phi$ is fixed and one step gradient descent is performed over $\theta$ for Equation \ref{eq:gan_loss1}. Then $\theta$ is fixed and one step gradient descent is performed over $\phi$ for Equation \ref{eq:gan_loss2}. The authors make the assumptions of the infinite capacity of $G$ and $D$. They also assume that $D$ is trained to convergence at every iteration.
\subsection{WGAN-GP}

In \cite{arjovsky01}, the authors introduce the issue of \emph{vanishing gradient}. If both $p_g$ and $p_d$ lie on different manifolds, the discriminator is easily able to achieve zero loss and the gradients w.r.t. $G$ vanish, leading to the vanishing gradient problem. The authors also prove that using Equation \ref{eq:gan_loss2} leads to mode collapse and unstable updates. Hence, in \cite{arjovskyWGAN}, the authors propose to use Wasserstein distance between the distributions. The loss is given as follows,
\begin{equation}
    \label{eq:wgan}
    \begin{split}
        \max_{\theta} \min_{\phi} \ V_{W}(D_{\theta},G_{\phi}) = & \mathbb{E}_{x\sim p_{d}(x)} [D_{\theta}(x)] - \\
        & \mathbb{E}_{z\sim p_{z}(z)} [D_{\theta}(G_{\phi}(z))]
    \end{split}{}
\end{equation}{}
The $D$ has to be $1$-Lipschitz w.r.t. $\theta$. In order to enforce that, the authors clamp $\theta$ to be within a specified range. Using this objective, the training is more stable and less sensitive to the optimization. 

In \cite{gularajani}, the authors show that clamping weights like in WGAN leads to the problem of exploding and vanishing gradients; hence they introduce a more elegant way for enforcing Lipschitz constraint on $D$. They introduce a gradient penalty term (GP) in Equation \ref{eq:wgan} to form the WGAN-GP loss as follows,

\begin{equation}
\label{eq:wgan_gp}
    \begin{aligned}
        V_{W}(D_{\theta},G_{\phi}) = \mathbb{E}_{x\sim p_{d}(x)} [D_{\theta}(x)] & - \mathbb{E}_{z\sim p_{z}(z)} [D_{\theta}(G_{\phi}(z))]  \\
        & + \lambda\mathbb{E}_{\tilde{x}\sim p_{\tilde{x}}}[(\parallel\nabla_{\tilde{x}}D(\tilde{x}) \parallel_{2} - 1)^2]
    \end{aligned}{}
\end{equation}{}

We obtain $ \tilde{x} \sim p_{\tilde{x}}$ by sampling uniformly along straight lines between pairs of points sampled from the data distribution $p_d$ and the generator distribution $p_{g}.$ The optimization method used is similar to GAN. Given the two different models used, we next state the measures that we use to evaluate quantitatively the performance of the samples generated.

\subsection{Measures to Evaluate GANs}
In general, it is not possible to compute how close $p_g$ is to $p_d$ quantitatively, given that GANs do not provide the distribution explicitly.
For the synthetic Gaussian data, we compute the empirical distance between $p_g$ and $p_d$ using the three measures, i) JSD  ii) \emph{Fr\'{e}chet Distance} (FD).
The definitions of these measures are as follows,
\begin{definition}[JSD]
The Jenson Shannon Divergence is a symmetric distance metric between the two distribution $p_d(x), p_g(x)$ given by,
\begin{equation}
    \label{eq:jsd}
    JSD(p_d\parallel p_g) = \frac{1}{2} KL \left(p_d \parallel \frac{p_d + p_g}{2}\right) + \frac{1}{2} KL \left(p_g \parallel \frac{p_d + p_g}{2} \right)
\end{equation}{}
where $KL$ is the Kl-divergence.
\end{definition}{}

\begin{definition}[FD]
Given $p_g$ and $p_d$ both multivariate continuous Gaussian distributions. The mean and variance of $p_g$ is $\mu_g, \Sigma_g$ and $p_d$ is $\mu_d, \Sigma_d$ respectively. The FD is then,
\begin{equation}
    \label{eq:fd}
    \parallel \mu_g - \mu_d \parallel^2_2  + Tr(\Sigma_g + \Sigma_d - 2 (\Sigma_g \Sigma_d)^{\frac{1}{2}})
\end{equation}{}
\end{definition}{}

 In our experiments on MNIST and CelebA datasets,  it is not possible to use the above measures; hence, we use the following standard performance measures,
\subsubsection{Inception Score (IS)} This is the most widely used score and was proposed by \cite{goodfellow16}. The pre-trained neural network named Inception Net is used. The score function defined measures the average distance between the label distribution $p(y|x)$ and marginal distribution $p(y)$. Here the $x$ is the image generated by $G$, and $y$ is the label given by the Inception Net. The distribution $p(y|x)$ needs to have less entropy, which indicates that the network can classify $x$ with high confidence hence more likely to be a realistic image. At the same time, $p(y)$ needs to have high entropy to indicate diversity in the samples generated. Hence, the higher the inception score, the better is the generator. 

\subsubsection{Fr\'{e}chect Inception Distance (FID)} Proposed by \cite{ttur} also uses a pre-trained Inception Net. The activations of the intermediate pooling layer serve as our feature embeddings. It is assumed that these embeddings follow a multivariate continuous Gaussian distribution. We pass multiple samples of $x \sim p_d$ and calculate the empirical mean and variance of their embeddings. Similarly, we sample $p_g$ and calculate the empirical mean and variance. The FD, given by Equation \ref{eq:fd}, is applied over the mean and variance of the two Gaussian embeddings. If the $p_g$ is close to $p_d$, the FD will be low. Hence lower the score, the better is the generator.

Apart from quantitative evaluation, we also provide the samples of images generated for each type of input noise, for visual comparison.
\section{Experiments with Input Noise for GANs}
\label{sec:expt}
In this section, we empirically study the effect of input noise on GANs. 
\subsection{Experimental Set-Up}
We used two different GANs in our experiments: (a) DCGAN and (b) WGAN-GP. We considered the following three data-sets for training and evaluation: \begin{itemize}
    \item Synthetic data-set: data generated with Gaussian distribution and/or mixture of Gaussian distribution
    \item The MNIST data-set
    \item CelebA (Celeb Faces Attribute A)
\end{itemize}

\noindent We studied the effect of dimension of input noise vector $z$, on FID and IS for all three data-sets and the both GANs. 
For quantitative comparison, we plot measures FID and IS against the dimension of the noise. For qualitative comparison, we looked at the original distributions (or images) and generated distributions (or images).

Below we describe the data-sets and the corresponding architectures used.

\subsection{Data-Sets and Architectures}
For a specific data-set, we have used a specific architecture. We keep the same architecture for both GAN and WGAN-GP losses as described by Equation \ref{eq:gan_loss1} and Equation \ref{eq:wgan} respectively.

\subsubsection*{\textbf{Synthetic Data-set}} We show results on 1-dimensional synthetic Gaussian data. This enables us to compare the plots for the distribution of the real data samples and generated samples. We study the effect of varying the variance of the Gaussian. We also vary the modes in the Gaussian (i.e., have two peaks) and visualize the problem of mode collapse). 

\emph{Architecture}: (vanilla GAN architecture) We use a simple feed-forward multi-layered perceptron having two hidden layers each. Both the generator and discriminator have ReLU activation.

\subsubsection*{\textbf{MNIST Data-set}} \cite{mnist} the data-set consists of $28 \times 28 $ dimensional black and white handwritten digits.

\emph{Architecture}: The discriminator has two convolution layers with 64 and 128 filters and stride 2 and leaky ReLU activation. There is a final dense layer to return a single value of probability. Generator has an architecture reverse to that of discriminator's but includes batch normalization layers after every transpose convolution layer. The activation used is ReLU.

\subsubsection*{\textbf{CelebA Data-set}} \cite{celeba} this data-set consists of more than 200K celebrity images, each of $128 \times 128$ resolution. We re-scale the images to $32 \times 32$ and $64 \times 64$, and train two different models for each of the resolutions. 

\emph{Architecture}: Here, the architecture is similar to that of MNIST, although the networks for $32 \times 32$ have 3 convolutional layers with 64, 128 and 256 filters and batch normalization layers even in the discriminator. For generating images of $64\times 64$, we include one extra convolutional layer which has 512 filters.

\subsection{Results for Synthetic Data-set}
\begin{figure}[!htb]
\centering
    \begin{subfigure}{.24\textwidth}
        \centering
        \includegraphics[width=.9\linewidth]{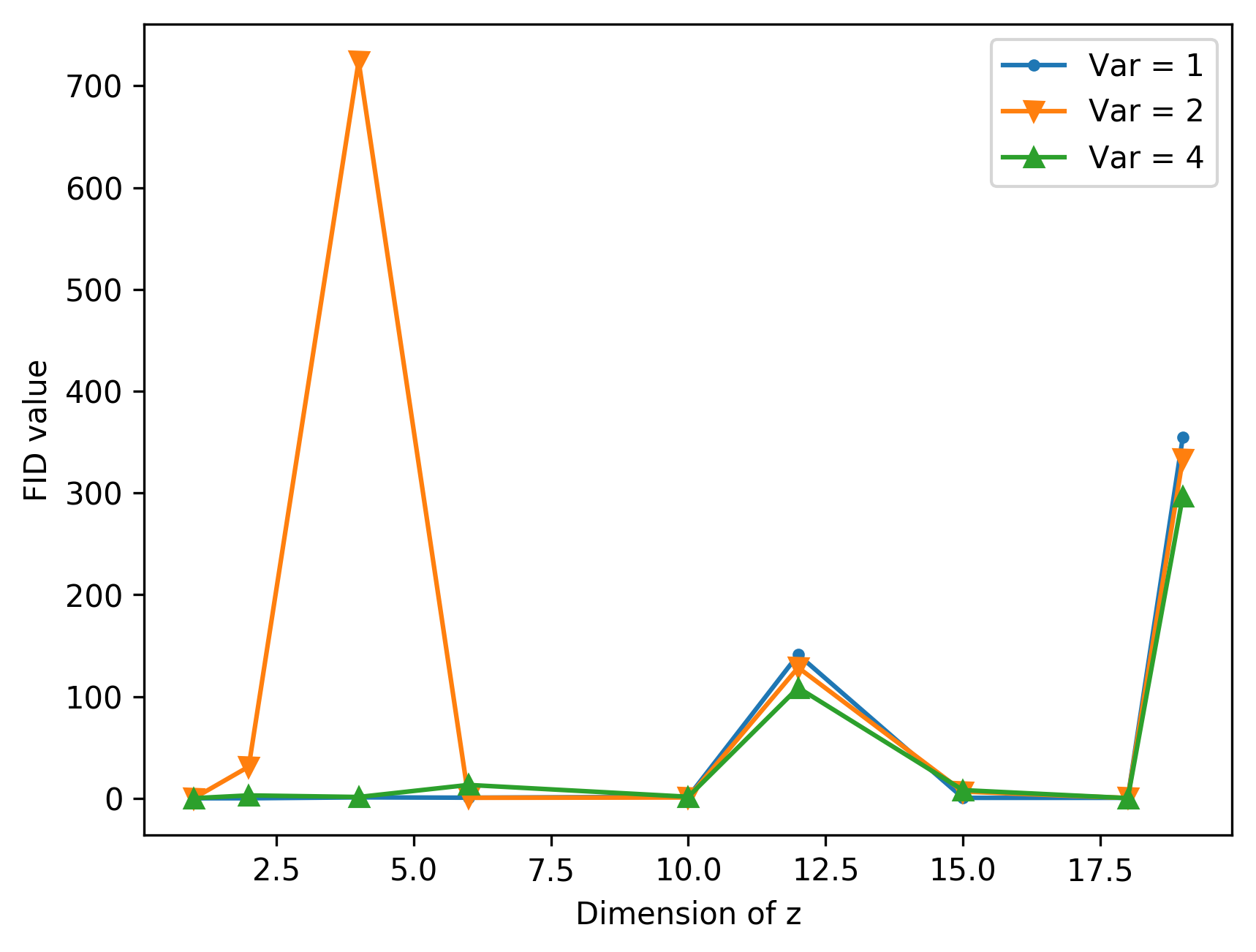}
        \caption{FD for Unimodal Gaussian Distribution in vanilla GAN}
        \label{fig:my_label}
    \end{subfigure} %
    \begin{subfigure}{.24\textwidth}
        \centering
        \includegraphics[width=.9\linewidth]{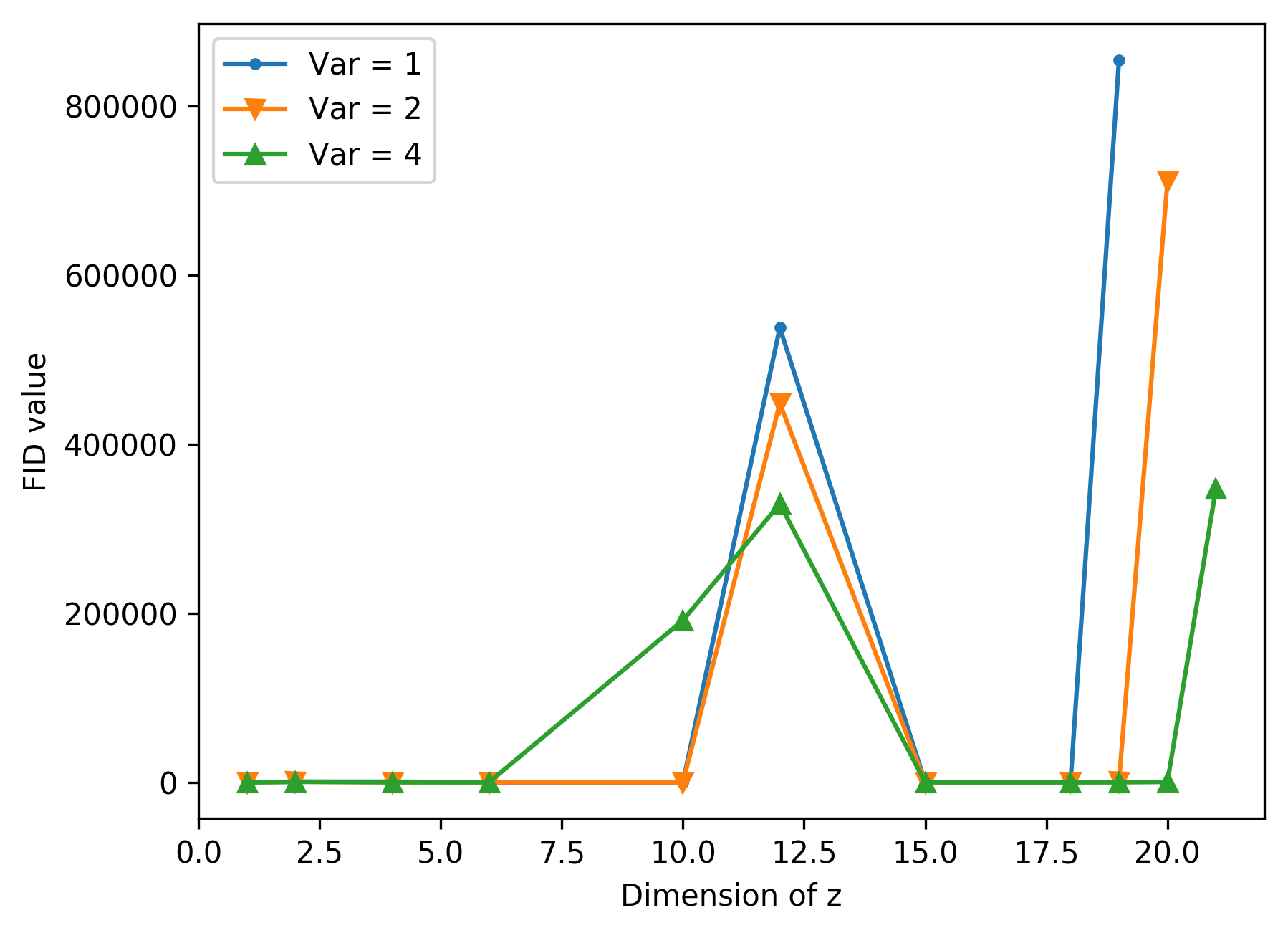}
        \caption{FD for Bimodal Gaussian Distribution in vanilla GAN}
        \label{fig:my_label}
    \end{subfigure}
        \begin{subfigure}{.24\textwidth}
        \centering
        \includegraphics[width=.9\linewidth]{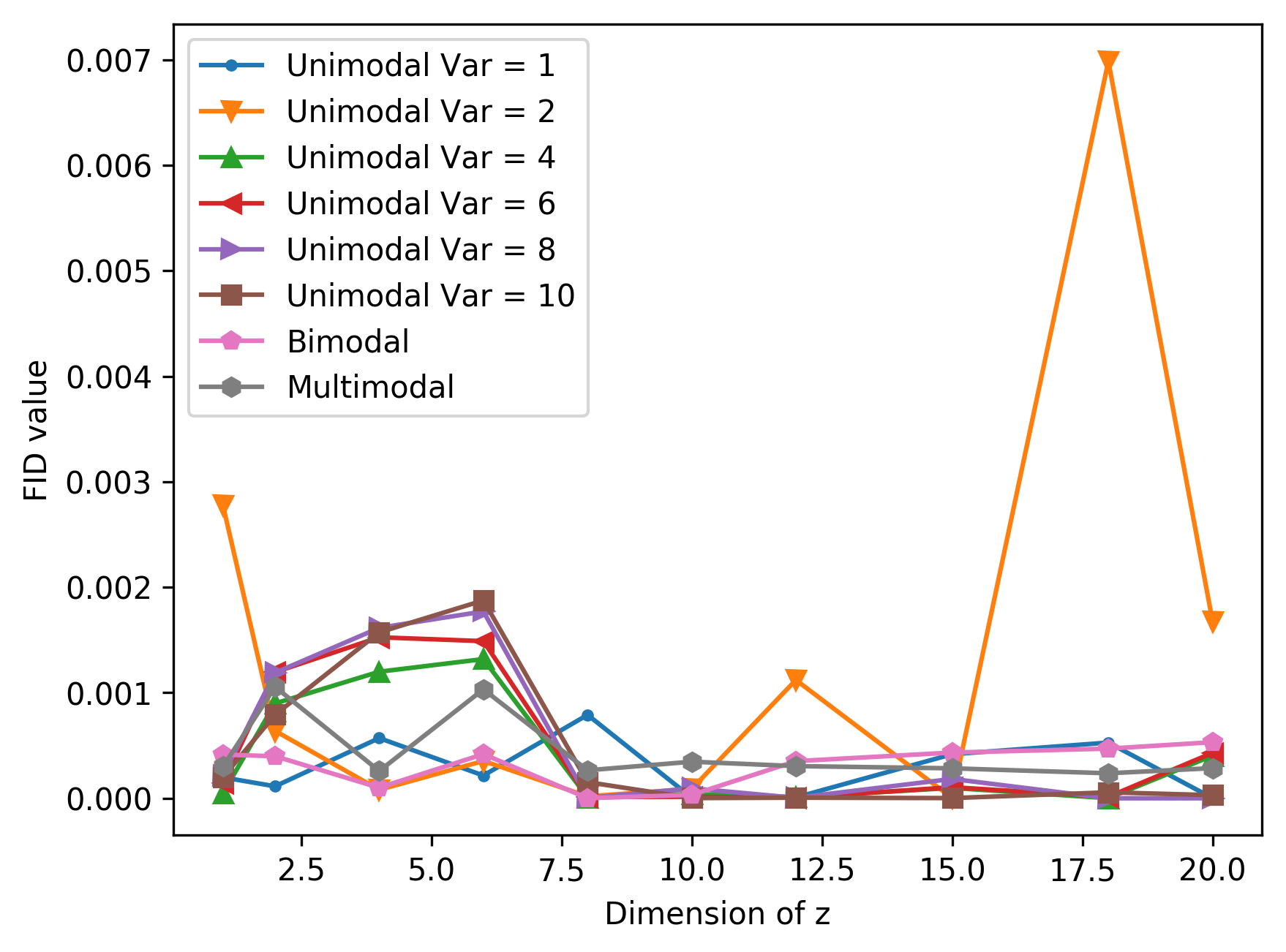}
        \caption{FD Plot for WGAN}
        \label{fig:my_label}
    \end{subfigure} %
    \begin{subfigure}{.24\textwidth}
        \centering
        \includegraphics[width=.9\linewidth]{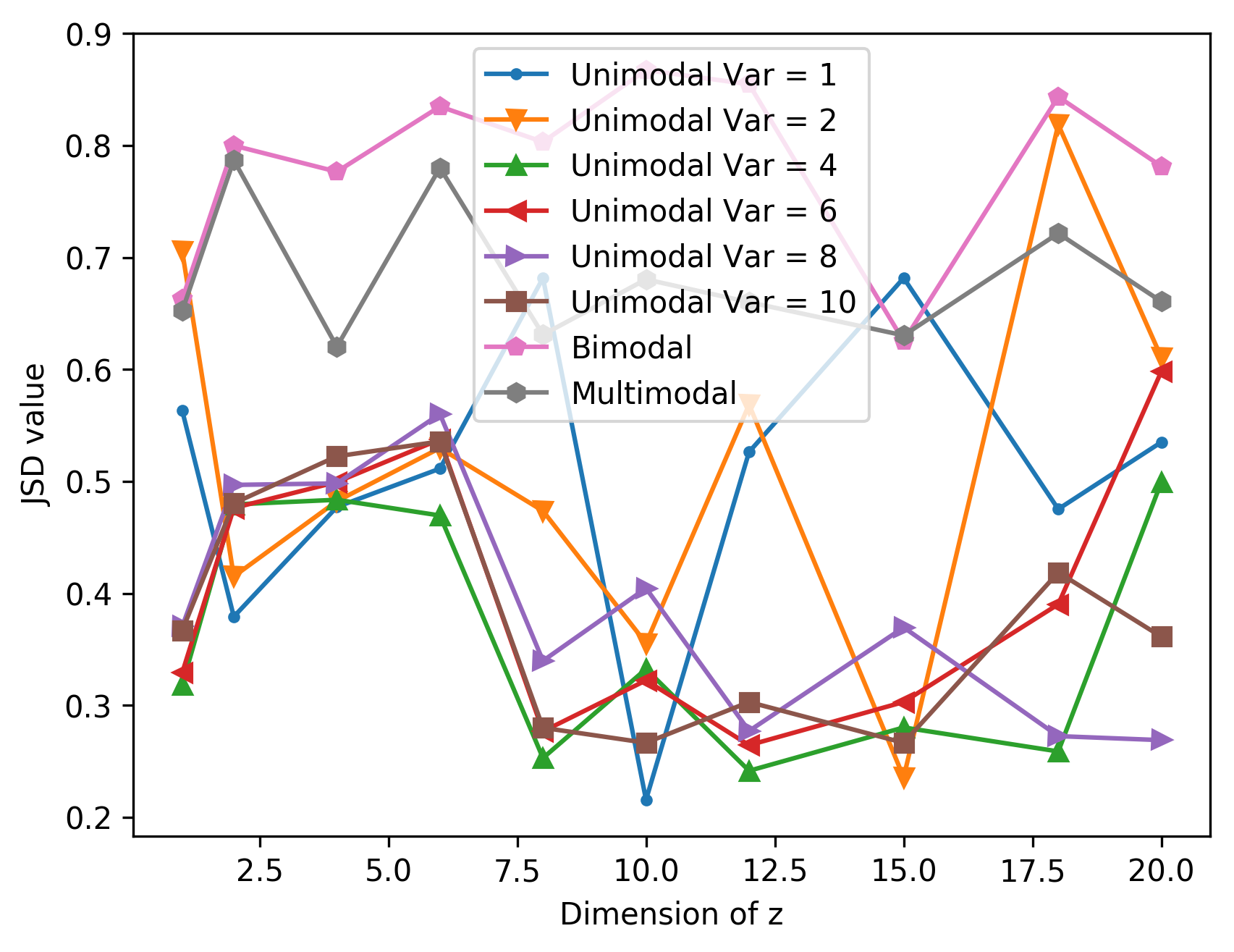}
        \caption{JSD Plot for WGAN}
        \label{fig:my_label}
    \end{subfigure}
\caption{Performance measure Plots}
\label{fig:gaussian1}
\end{figure}

\begin{figure}[!htb]
\centering
    \begin{subfigure}{.24\textwidth}
        \centering
        \includegraphics[width=.9\linewidth]{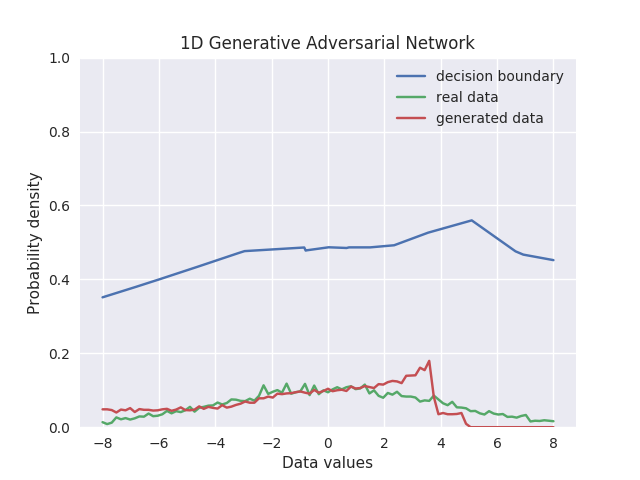}
        \caption{Unimodal Gaussian Distribution with Variance = 4 and dimension of noise = 10}
        \label{fig:my_label}
    \end{subfigure} %
    \begin{subfigure}{.24\textwidth}
        \centering
        \includegraphics[width=.9\linewidth]{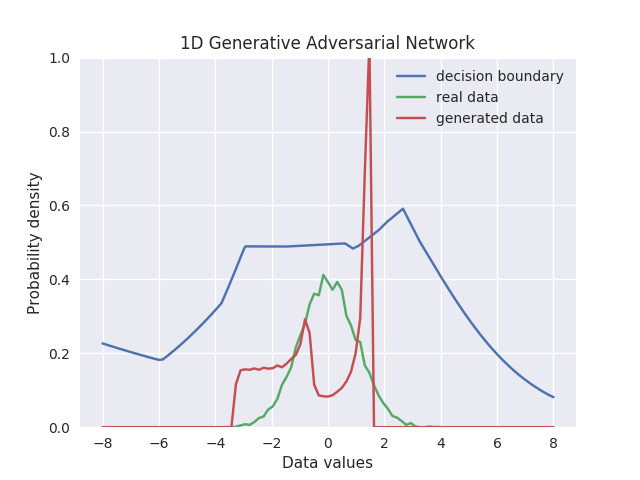}
        \caption{Unimodal Gaussian Distribution with Variance = 1 and dimension of noise = 18}
        \label{fig:my_label}
    \end{subfigure} %
    \begin{subfigure}{.24\textwidth}
        \centering
        \includegraphics[width=.9\linewidth]{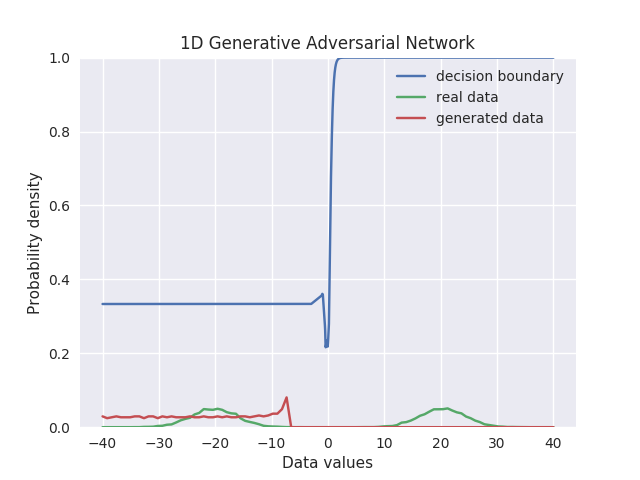}
        \caption{Bimodal Gaussian Distribution with dimension of noise = 10}
        \label{fig:my_label}
    \end{subfigure} %
    \begin{subfigure}{.24\textwidth}
        \centering
        \includegraphics[width=.9\linewidth]{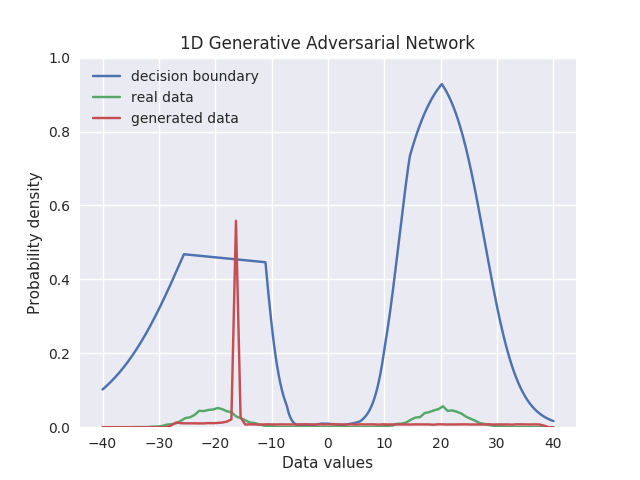}
        \caption{Bimodal Gaussian Distribution with dimension of noise = 19}
        \label{fig:my_label}
    \end{subfigure} %
\caption{Distributions generated using vanilla GAN}
\label{fig:gaussian2}
\end{figure}

In Figure \ref{fig:gaussian1} (a) and (b), we present the FD values (Equation \ref{eq:fd}) between the real and generated Gaussian data, i) the data has single mode and ii) bimodal data. It is observed that the variance in the data does not effect the trend in performance. The FD values stay low till the dimension of input noise is 10. Increasing the dimension only worsens the performance. And increasing it further from 20, for a fixed generator and discriminator architecture blows up the distance.
The problem of mode collapse is also evident as we see the FD values for bimodal is greater than the values in the unimodal case.
From Figure \ref{fig:gaussian2}, we can visually observe that the real data and generated data distributions are nearby with input noise dimension is 10. The distributions are far apart when the dimension is increased to around 20.

\begin{figure}[!htb]
\centering
    \begin{subfigure}{.24\textwidth}
        \centering
        \includegraphics[width=.9\linewidth]{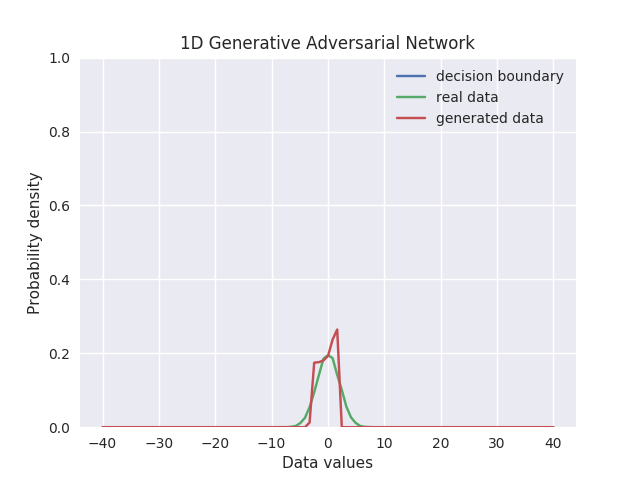}
        \caption{Unimodal Gaussian Distribution with Variance = 2 and dimension of noise = 10}
        \label{fig:my_label}
    \end{subfigure} %
    \begin{subfigure}{.24\textwidth}
        \centering
        \includegraphics[width=.9\linewidth]{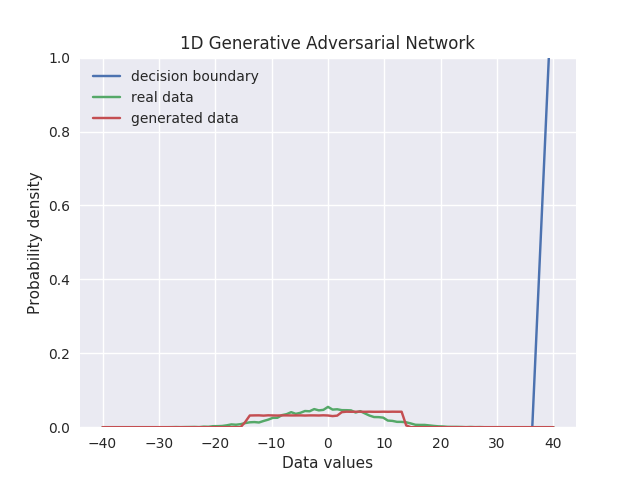}
        \caption{Unimodal Gaussian Distribution with Variance = 8 and dimension of noise = 20}
        \label{fig:my_label}
    \end{subfigure} %
     \begin{subfigure}{.24\textwidth}
        \centering
        \includegraphics[width=.9\linewidth]{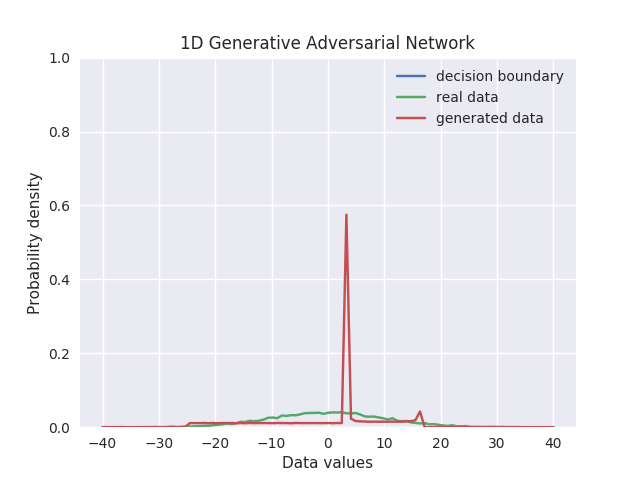}
        \caption{Unimodal Gaussian Distribution with Variance = 10 and dimension of noise = 6}
        \label{fig:my_label}
    \end{subfigure} %
    \begin{subfigure}{.24\textwidth}
        \centering
        \includegraphics[width=.9\linewidth]{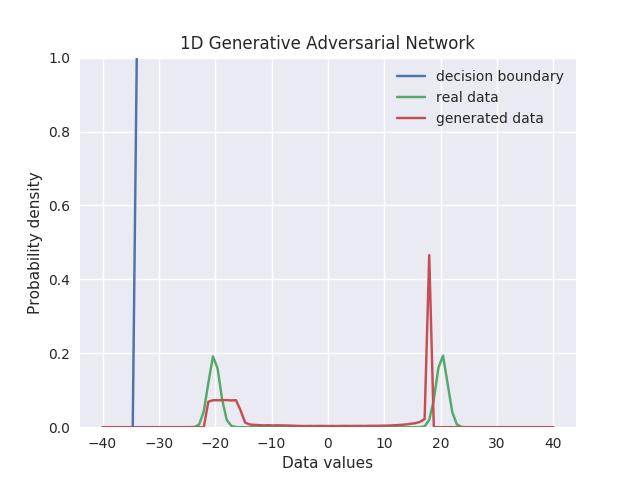}
        \caption{Bimodal Gaussian Distribution with Variance = 4 and dimension of noise = 10}
        \label{fig:my_label}
    \end{subfigure} %
\caption{Distributions generated using WGAN-GP}
\label{fig:gaussian4}
\end{figure}

In Figure \ref{fig:gaussian1} (c) and (d), we compare various performance measures, FD and JSD for WGAN-GP. We find that, there is no particular trend although there is small difference in the measures as the dimension increases and for all the variance and modes. We also observe that, the values are smaller than GAN. Hence, we conclude that WGAN performs better than the normal GAN on this data-set. This is also visible from the distribution plots in Figure \ref{fig:gaussian4}. The problem of mode collapse is still not completely overcome even in WGAN-GP.

\begin{figure}[!htb]
\centering
    \begin{subfigure}{.24\textwidth}
        \centering
        \includegraphics[width=.9\linewidth]{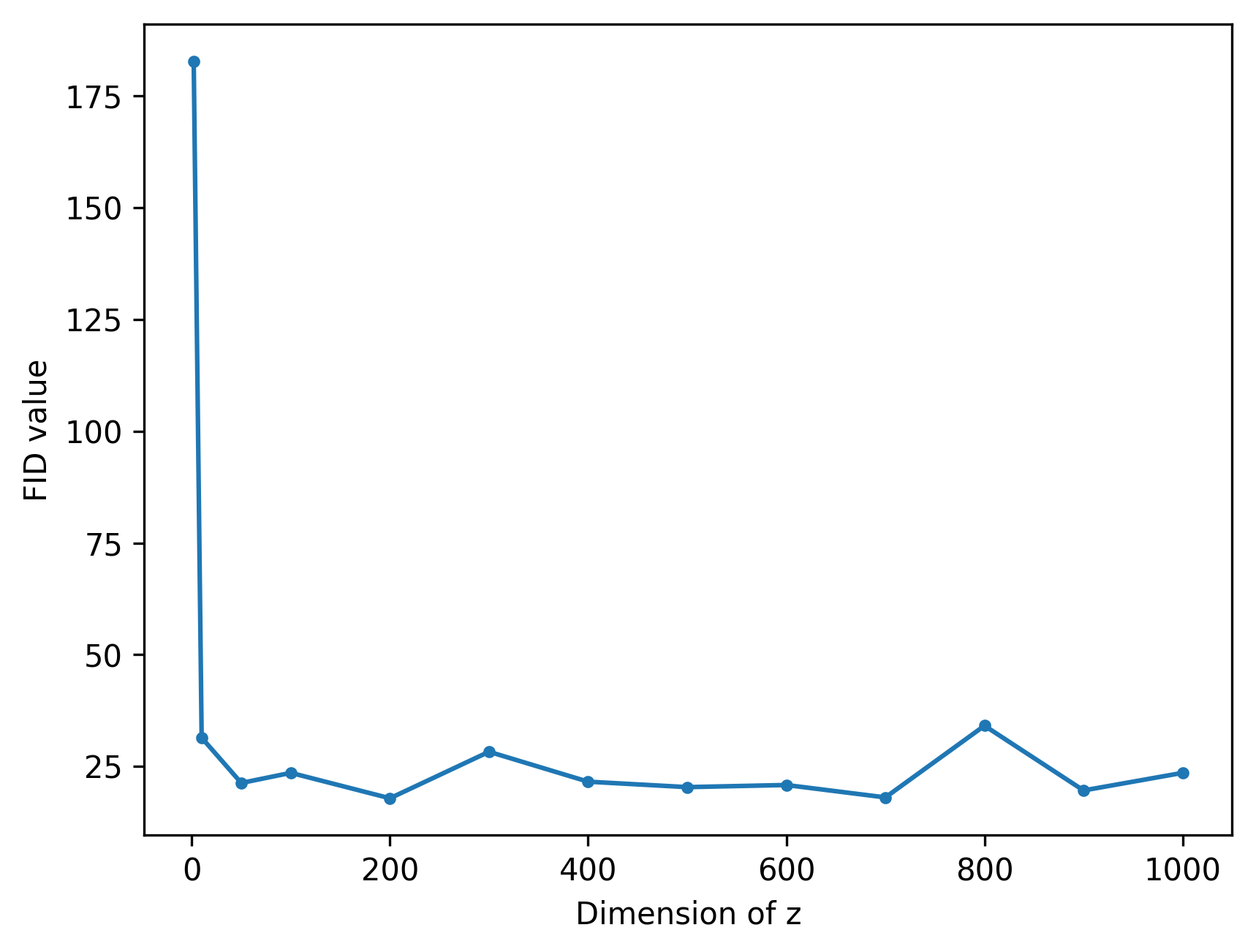}
        \caption{FID Plot for DCGAN}
        \label{fig:my_label}
    \end{subfigure} %
    \begin{subfigure}{.24\textwidth}
        \centering
        \includegraphics[width=.9\linewidth]{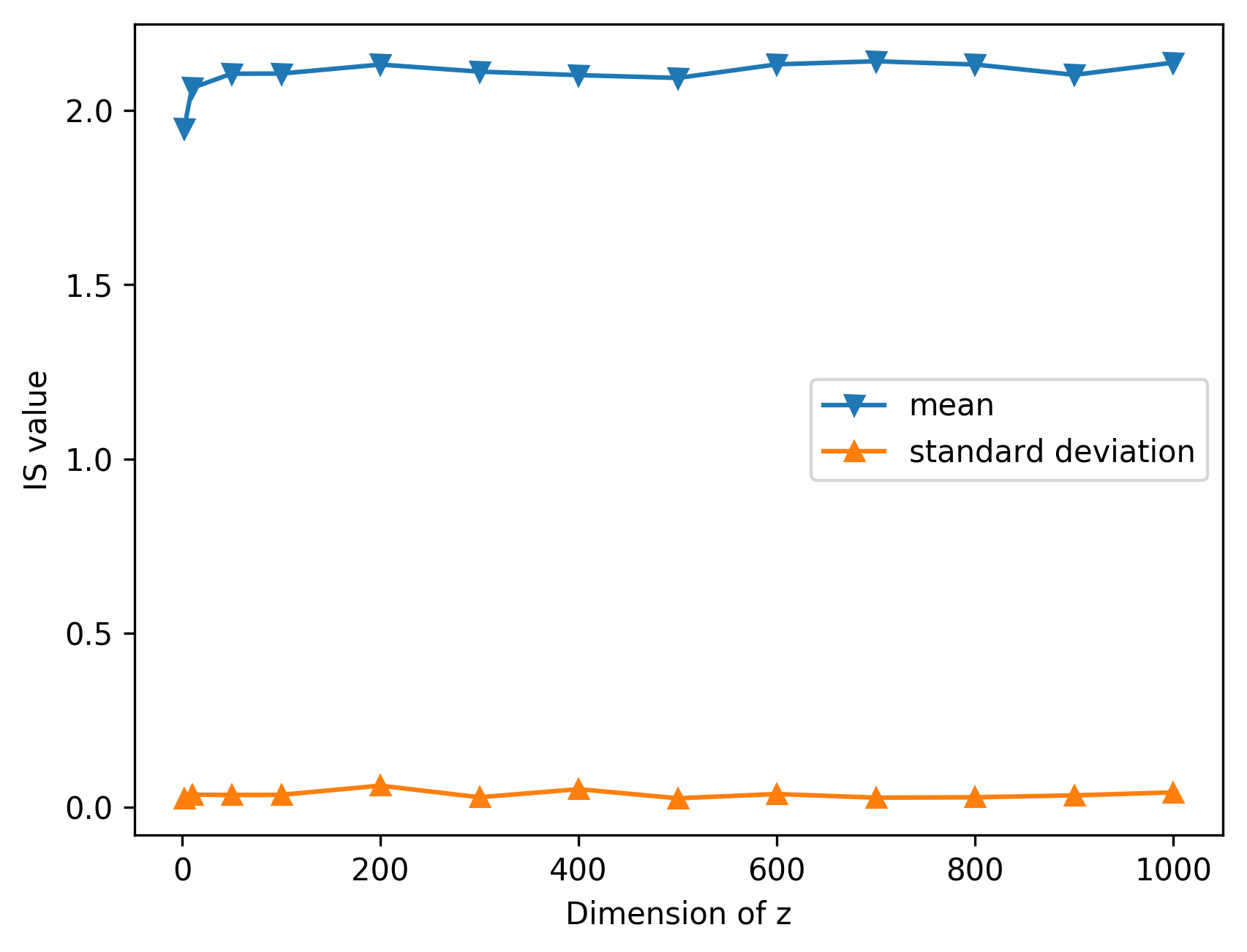}
        \caption{IS Plot for DCGAN}
        \label{fig:my_label}
    \end{subfigure}
    \begin{subfigure}{.24\textwidth}
        \centering
        \includegraphics[width=.9\linewidth]{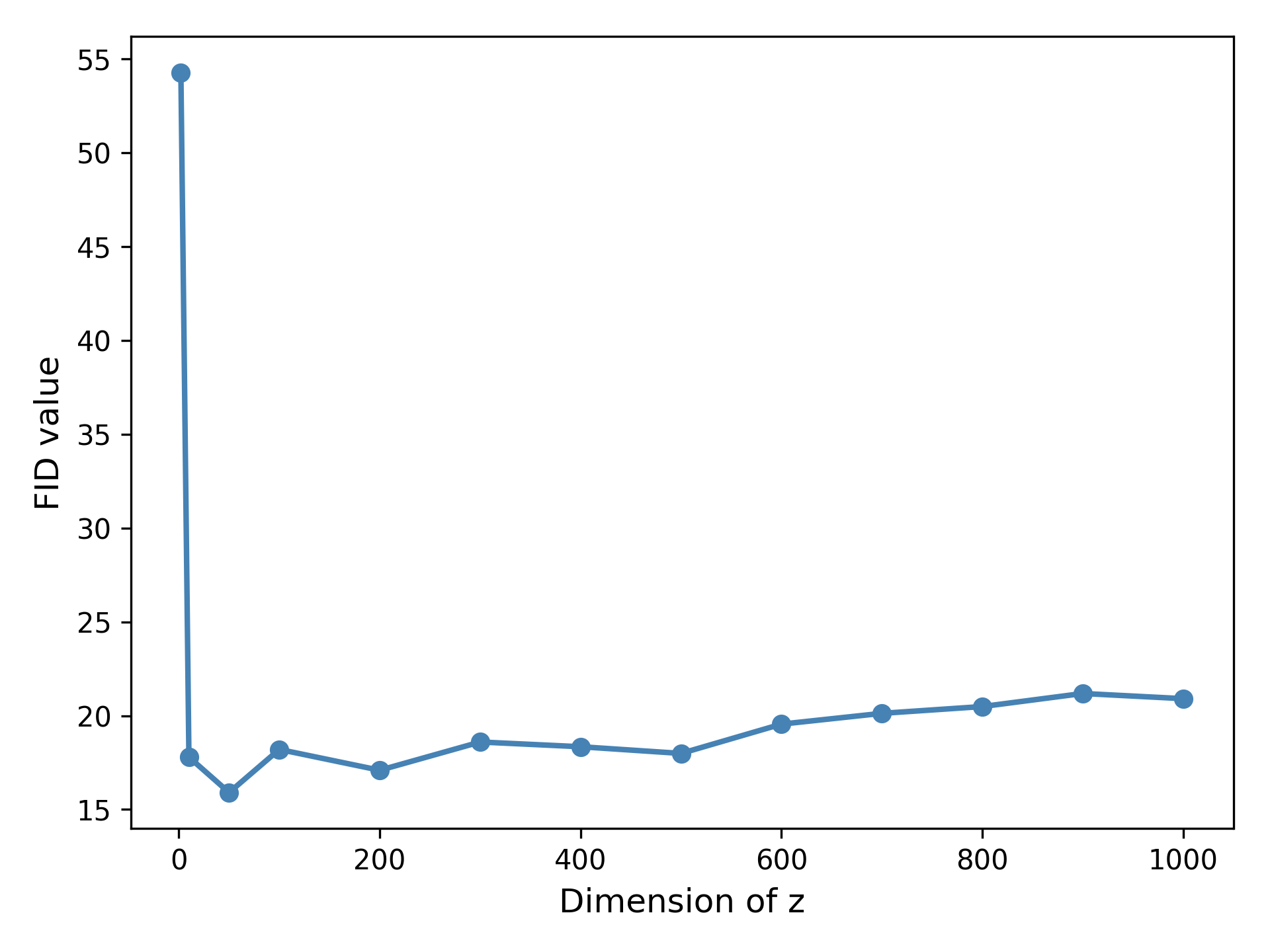}
        \caption{FID Plot for WGAN-GP}
        \label{fig:my_label}
    \end{subfigure} %
    \begin{subfigure}{.24\textwidth}
        \centering
        \includegraphics[width=.9\linewidth]{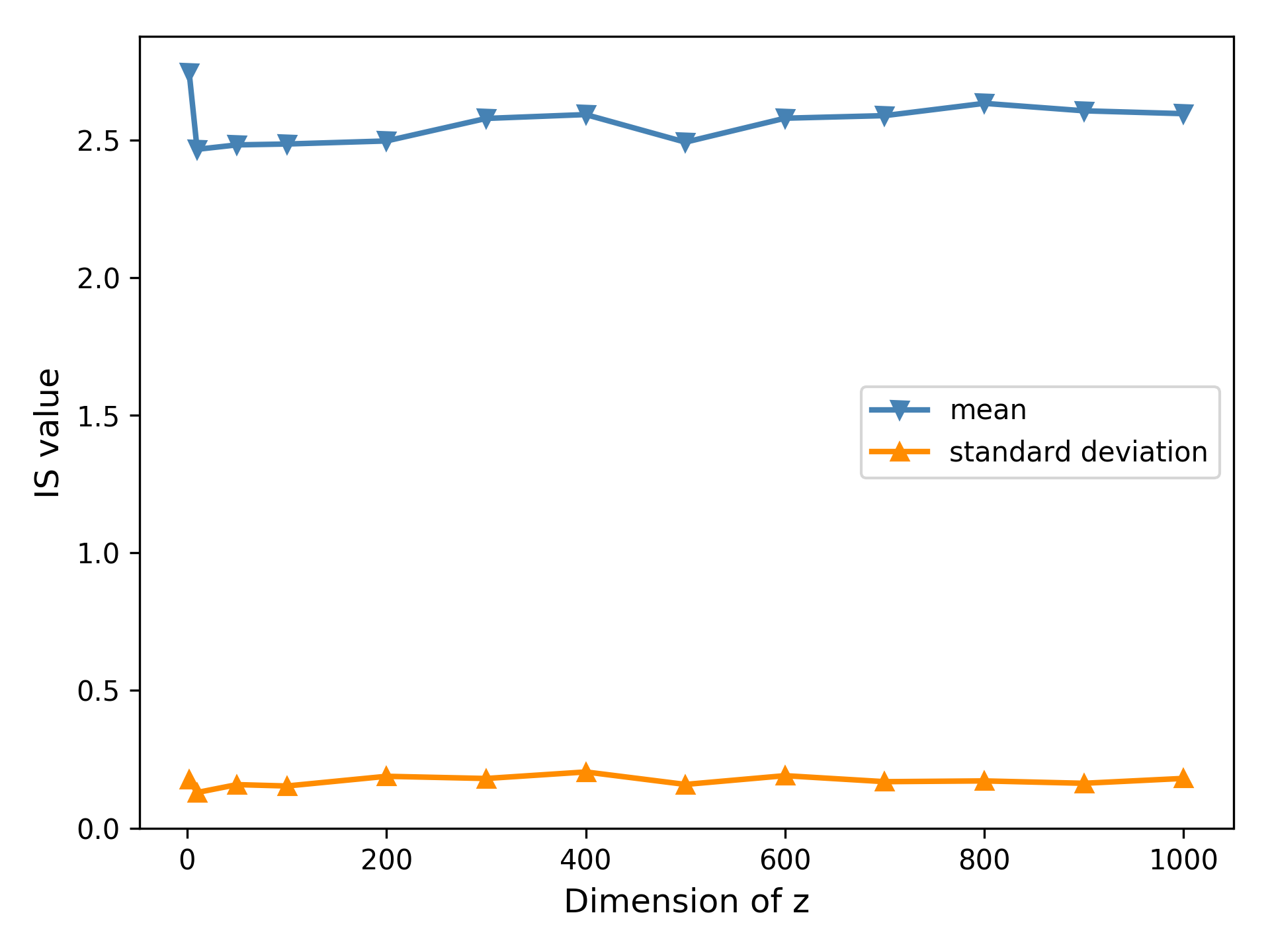}
        \caption{IS Plot for WGAN-GP}
        \label{fig:my_label}
    \end{subfigure}
\caption{Performance Measure Plots}
\label{fig:mnist1}
\end{figure}

\begin{figure}[!htb]
\centering
    \begin{subfigure}{.24\textwidth}
        \centering
        \includegraphics[width=.9\linewidth]{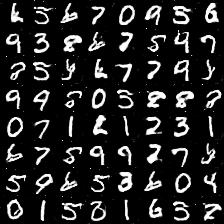}
        \caption{Dimension of noise = 2}
        \label{fig:my_label}
    \end{subfigure} %
    \begin{subfigure}{.24\textwidth}
        \centering
        \includegraphics[width=.9\linewidth]{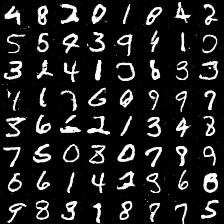}
        \caption{Dimension of noise = 10}
        \label{fig:my_label}
    \end{subfigure} %
    \begin{subfigure}{.24\textwidth}
        \centering
        \includegraphics[width=.9\linewidth]{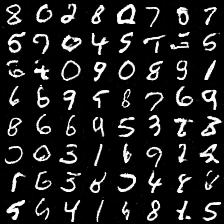}
        \caption{Dimension of noise = 100}
        \label{fig:my_label}
    \end{subfigure} %
    \begin{subfigure}{.24\textwidth}
        \centering
        \includegraphics[width=.9\linewidth]{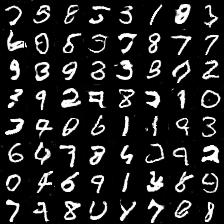}
        \caption{Dimension of noise = 1000}
        \label{fig:my_label}
    \end{subfigure} %
\caption{Images Generated by DCGAN}
\label{fig:mnist2}
\end{figure}

\begin{figure}[!htb]
\centering
    \begin{subfigure}{.24\textwidth}
        \centering
        \includegraphics[width=.9\linewidth]{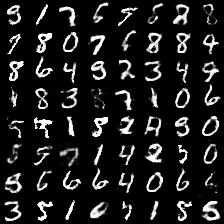}
        \caption{Dimension of noise = 2}
        \label{fig:my_label}
    \end{subfigure} %
    \begin{subfigure}{.24\textwidth}
        \centering
        \includegraphics[width=.9\linewidth]{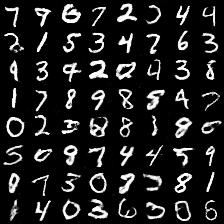}
        \caption{Dimension of noise = 10}
        \label{fig:my_label}
    \end{subfigure} %
    \begin{subfigure}{.24\textwidth}
        \centering
        \includegraphics[width=.9\linewidth]{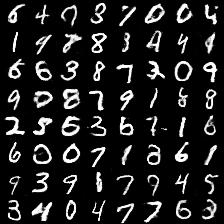}
        \caption{Dimension of noise = 100}
        \label{fig:my_label}
    \end{subfigure} %
    \begin{subfigure}{.24\textwidth}
        \centering
        \includegraphics[width=.9\linewidth]{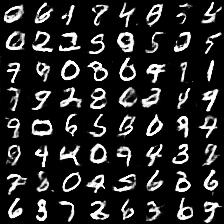}
        \caption{Dimension of noise = 1000}
        \label{fig:my_label}
    \end{subfigure} %
\caption{Images Generated by WGAN-GP}
\label{fig:mnist3}
\end{figure}

\subsection{Results for MNIST}
In Figure \ref{fig:mnist1} (a) and (c), we compare the FID values for different dimensions of input noise for DCGAN and WGAN-GP. We fix the architecture of the generator and the discriminator and we find that the FID scores are very high for  noise dimension 2 but do not change much for higher dimensions till 1000. At the same time, we find the FID values for WGAN-GP are much better than DCGAN. In Figure \ref{fig:mnist1} (b) and (d), we plot the IS values which are evaluated for batches of samples and the mean and the variance of the IS values across the batches is plotted for both DCGAN and WGAN-GP. We observe a similar trend as the FID. 

We visually compare the results in Figures \ref{fig:mnist2}, \ref{fig:mnist3}. We find that results are bad when noise dimension is 2 compared to the other dimensions. WGAN-GP performs worse compared to DCGAN at lower-dimensional input noise.
Hence we conclude that having dimension of noise as 10 is sufficient for good performance. 

\subsection{Results for CelebA 32}
\begin{figure}[!htb]
\centering
    \begin{subfigure}{.24\textwidth}
        \centering
        \includegraphics[width=.9\linewidth]{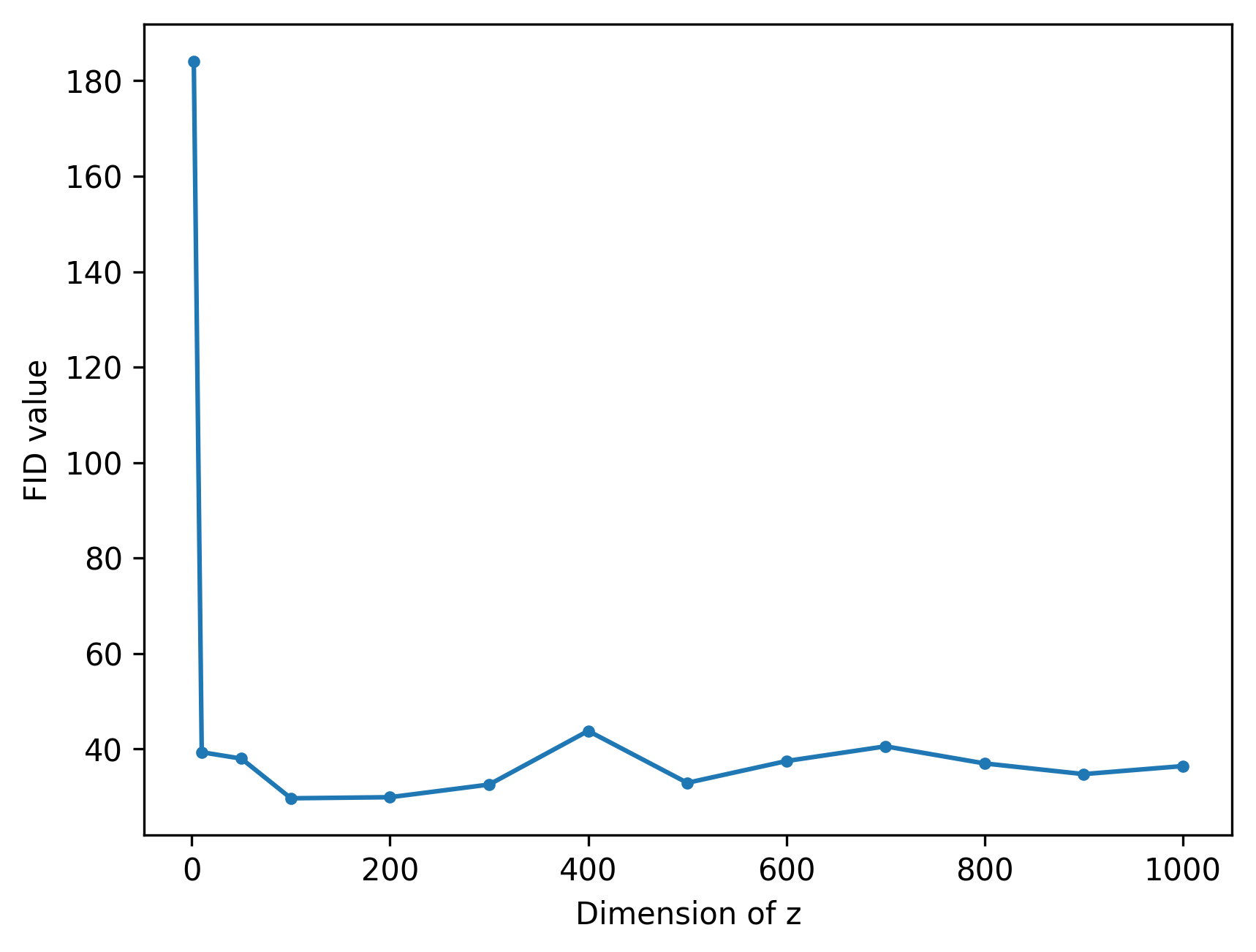}
        \caption{FID Plot}
        \label{fig:my_label}
    \end{subfigure} %
    \begin{subfigure}{.24\textwidth}
        \centering
        \includegraphics[width=.9\linewidth]{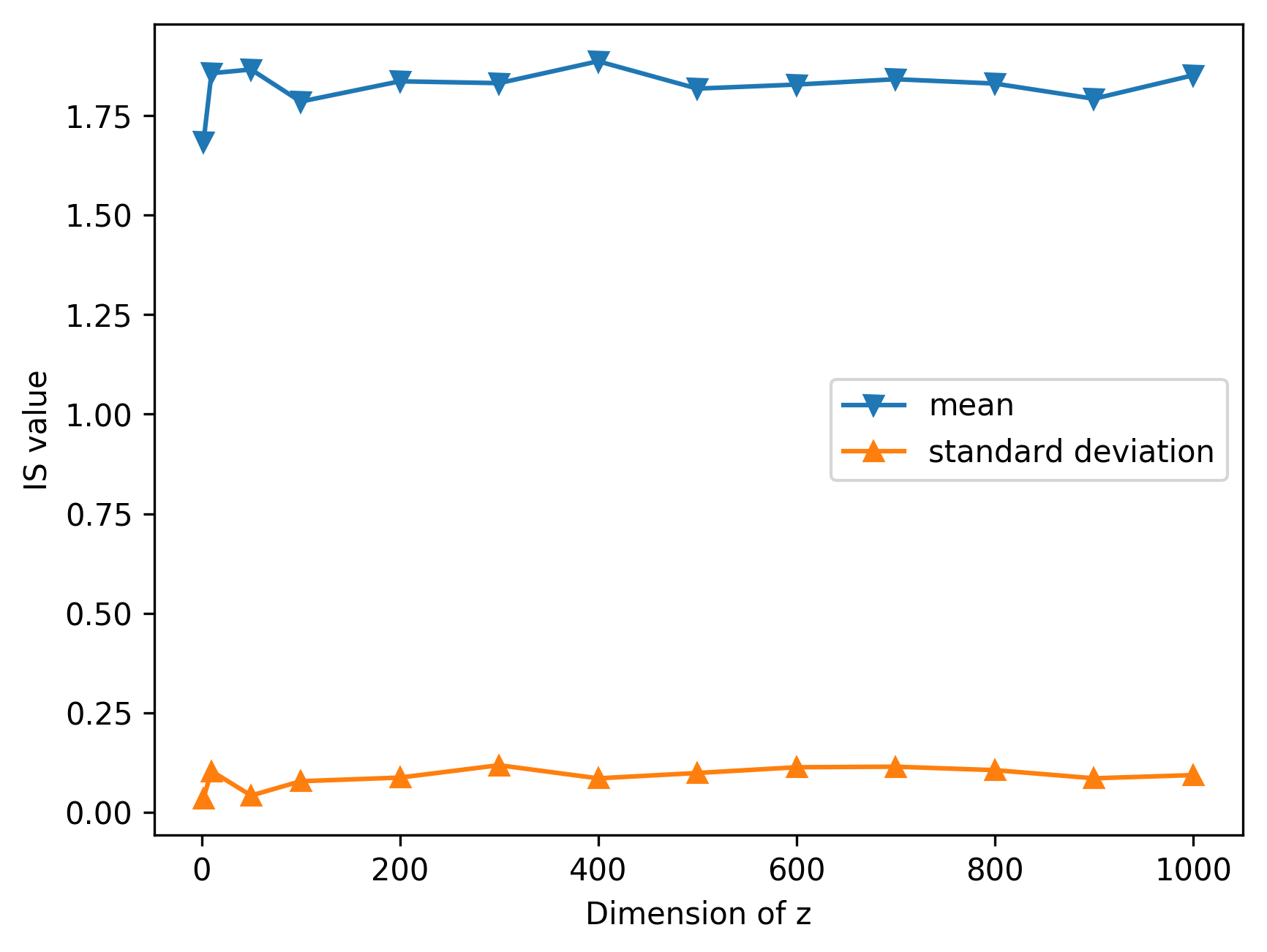}
        \caption{IS Plot}
        \label{fig:my_label}
    \end{subfigure}
    \begin{subfigure}{.24\textwidth}
        \centering
        \includegraphics[width=.9\linewidth]{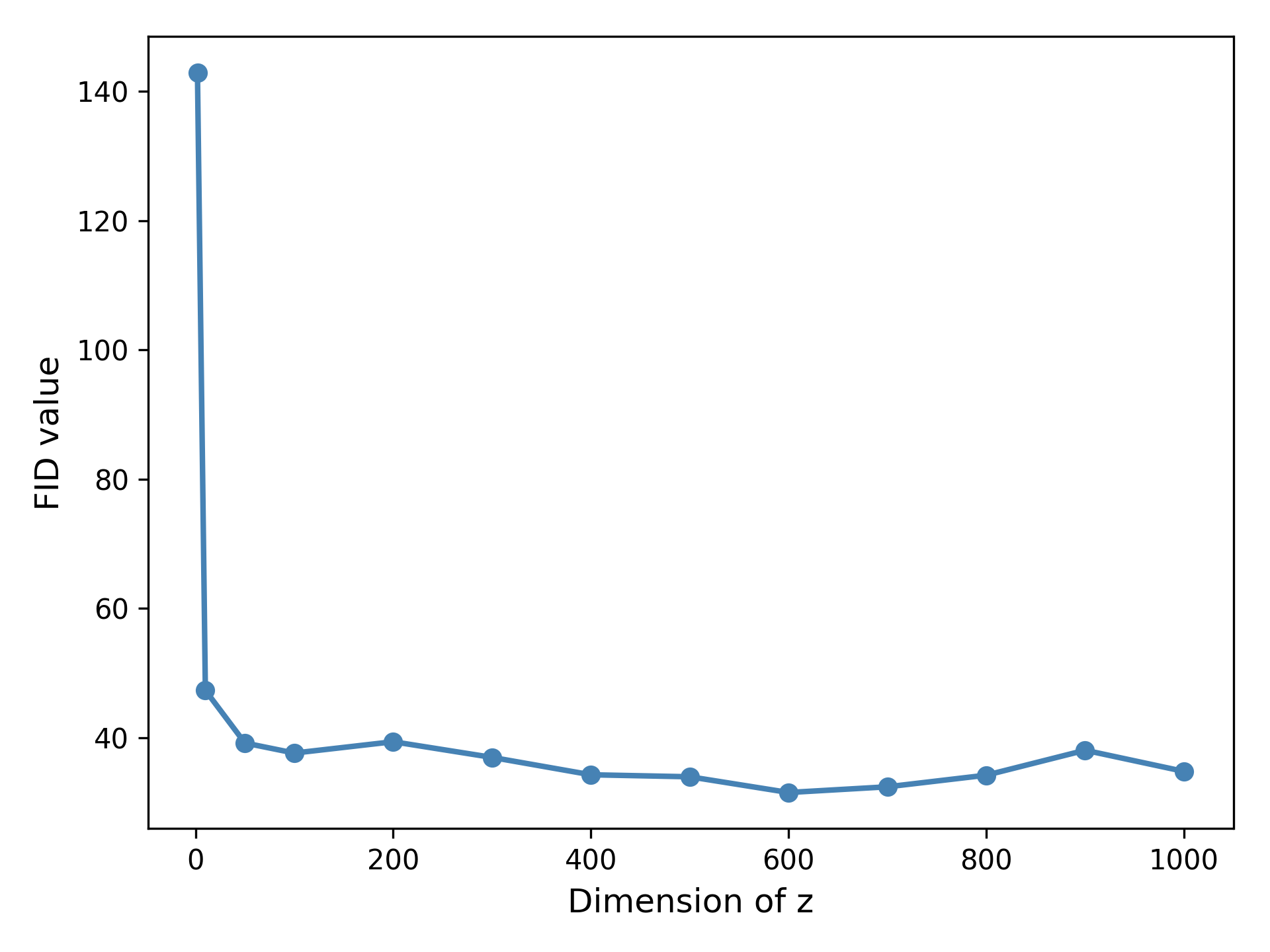}
        \caption{FID Plot}
        \label{fig:my_label}
    \end{subfigure} %
    \begin{subfigure}{.24\textwidth}
        \centering
        \includegraphics[width=.9\linewidth]{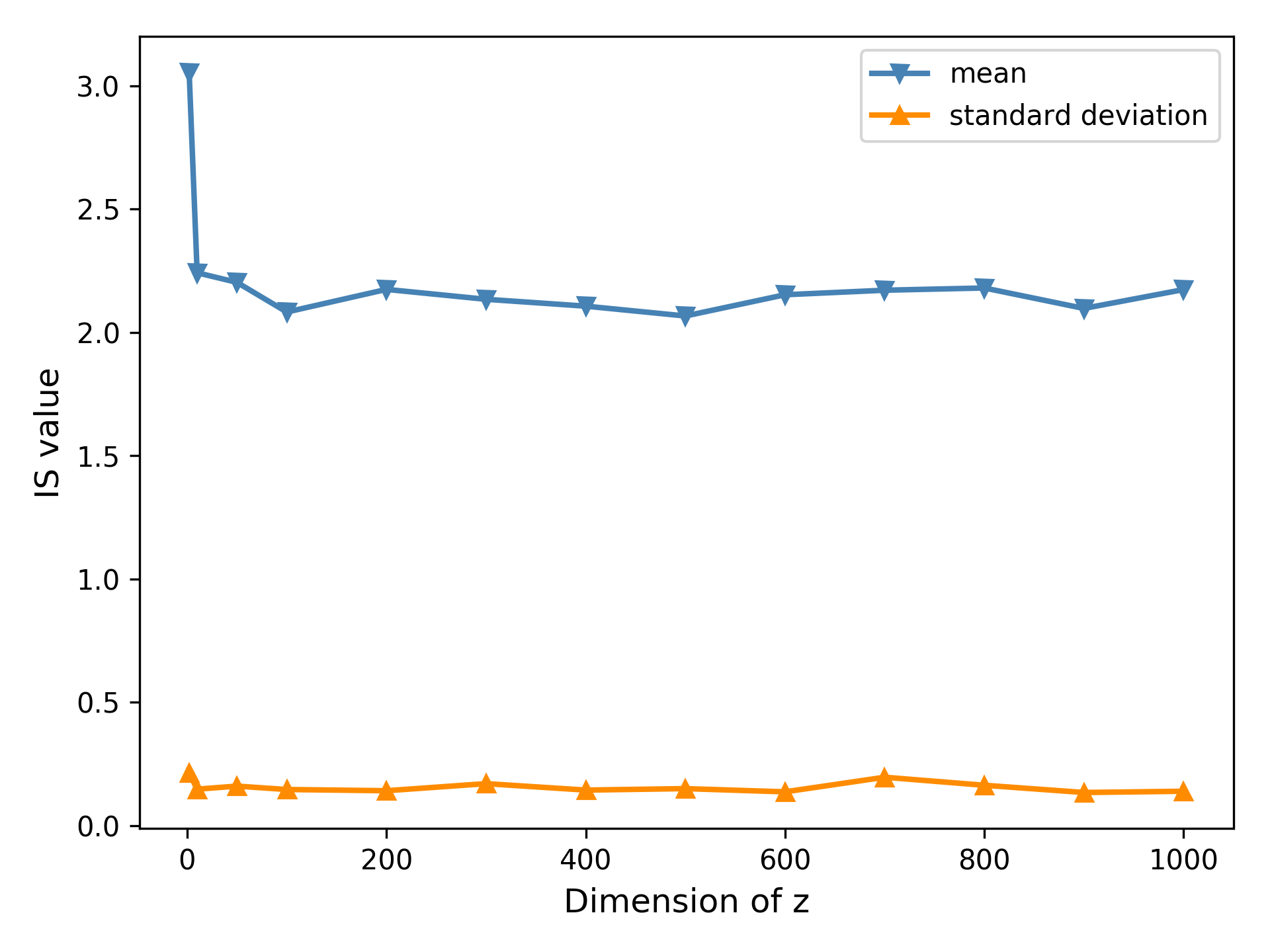}
        \caption{IS Plot}
        \label{fig:my_label}
    \end{subfigure}
\caption{Performance Measure Plots for 32x32 CelebA}
\label{fig:celeba1}
\end{figure}

\begin{figure}[!htb]
\centering
    \begin{subfigure}{.24\textwidth}
        \centering
        \includegraphics[width=.9\linewidth]{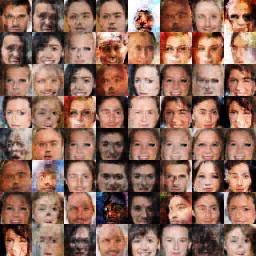}
        \caption{Dimension of noise = 2}
        \label{fig:my_label}
    \end{subfigure} %
    \begin{subfigure}{.24\textwidth}
        \centering
        \includegraphics[width=.9\linewidth]{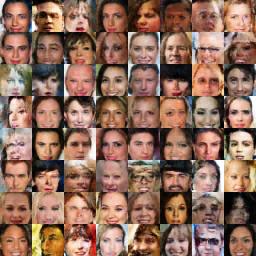}
        \caption{Dimension of noise = 10}
        \label{fig:my_label}
    \end{subfigure} %
    \begin{subfigure}{.24\textwidth}
        \centering
        \includegraphics[width=.9\linewidth]{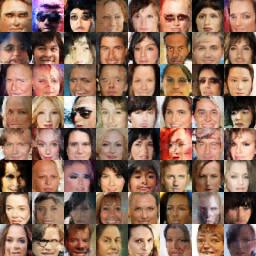}
        \caption{Dimension of noise = 100}
        \label{fig:my_label}
    \end{subfigure} %
    \begin{subfigure}{.24\textwidth}
        \centering
        \includegraphics[width=.9\linewidth]{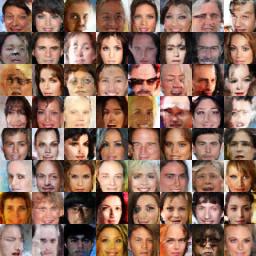}
        \caption{Dimension of noise = 900}
        \label{fig:my_label}
    \end{subfigure} %
\caption{Images Generated by DCGAN for 32x32 CelebA}
\label{fig:celeba2}
\end{figure}

\begin{figure}[!htb]
\centering
    \begin{subfigure}{.24\textwidth}
        \centering
        \includegraphics[width=.9\linewidth]{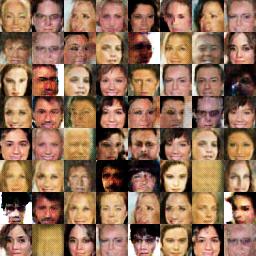}
        \caption{Dimension of noise = 2}
        \label{fig:my_label}
    \end{subfigure} %
    \begin{subfigure}{.24\textwidth}
        \centering
        \includegraphics[width=.9\linewidth]{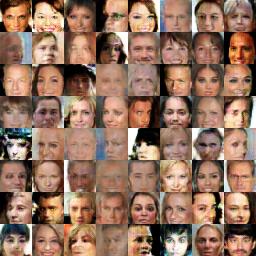}
        \caption{Dimension of noise = 10}
        \label{fig:my_label}
    \end{subfigure} %
    \begin{subfigure}{.24\textwidth}
        \centering
        \includegraphics[width=.9\linewidth]{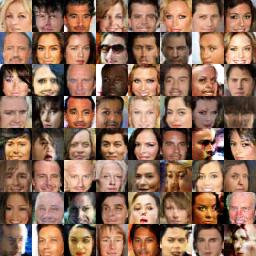}
        \caption{Dimension of noise = 100}
        \label{fig:my_label}
    \end{subfigure} %
    \begin{subfigure}{.24\textwidth}
        \centering
        \includegraphics[width=.9\linewidth]{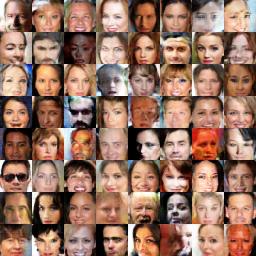}
        \caption{Dimension of noise = 900}
        \label{fig:my_label}
    \end{subfigure} %
\caption{Images Generated by WGAN-GP for 32x32 CelebA}
\label{fig:celeba3}
\end{figure}

In Figure \ref{fig:celeba1} (a) and (c), we compare the FID and IS values for generating CelebA images for different dimensions of input noise for DCGAN and WGAN-GP. We fix the architecture of the generator and the discriminator and we find that the FID scores are very high noise dimension 2 and then reduce drastically. Further increasing the dimension does not effect the FID values much. Both WGAN and DCGAN perform almost equally. Although WGAN is slightly better. In Figure \ref{fig:celeba1} (b) and (d), we plot the IS values which are evaluated for batches of samples and the mean and the variance of the IS values across the batches is plotted for both DCGAN and WGAN-GP. Figures \ref{fig:celeba2}, \ref{fig:celeba3}. We find that results when noise dimension is 2 is bad compared to the other dimensions. For dimension 2 and 10 the images generated using WGAN lack clarity while that of DCGAN lack variety. For dimension 100 and 900, visually their performance is similar.

\subsection{Results for CelebA 64}
\begin{figure}[!htb]
\centering
    \begin{subfigure}{.24\textwidth}
        \centering
        \includegraphics[width=.9\linewidth]{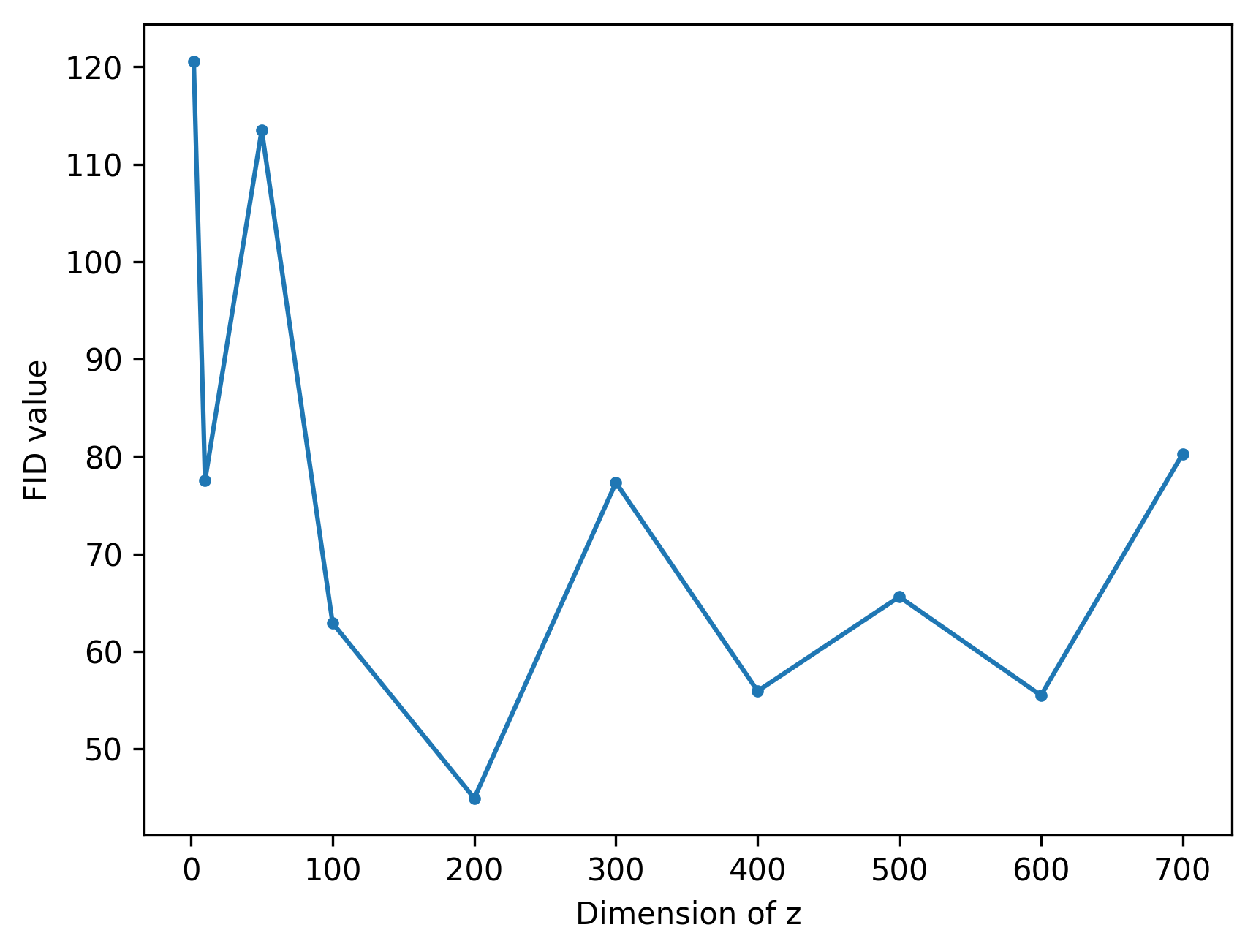}
        \caption{FID Plot}
        \label{fig:my_label fo}
    \end{subfigure} %
    \begin{subfigure}{.24\textwidth}
        \centering
        \includegraphics[width=.9\linewidth]{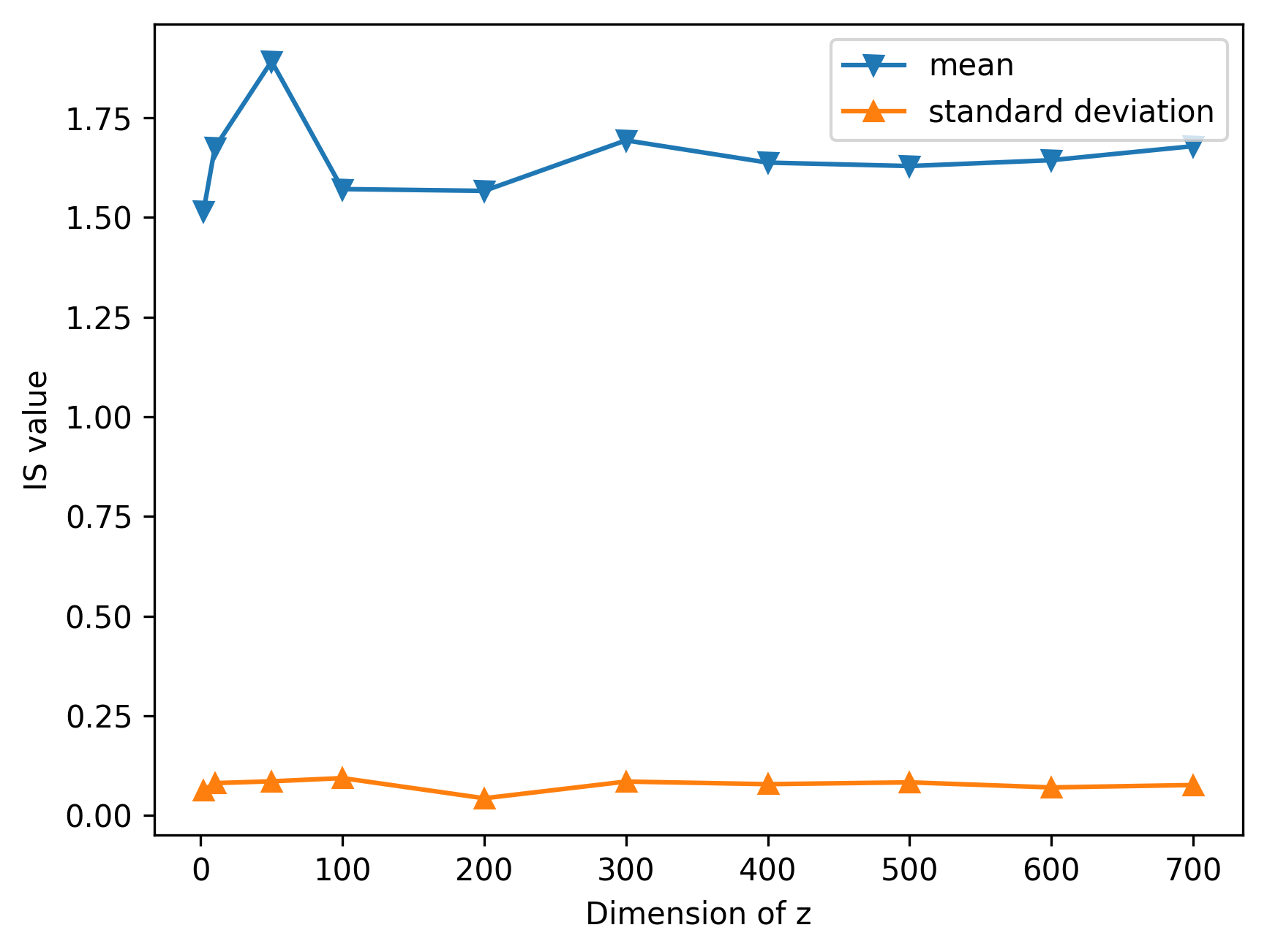}
        \caption{IS Plot}
        \label{fig:my_label}
    \end{subfigure}
    \begin{subfigure}{.24\textwidth}
        \centering
        \includegraphics[width=.9\linewidth]{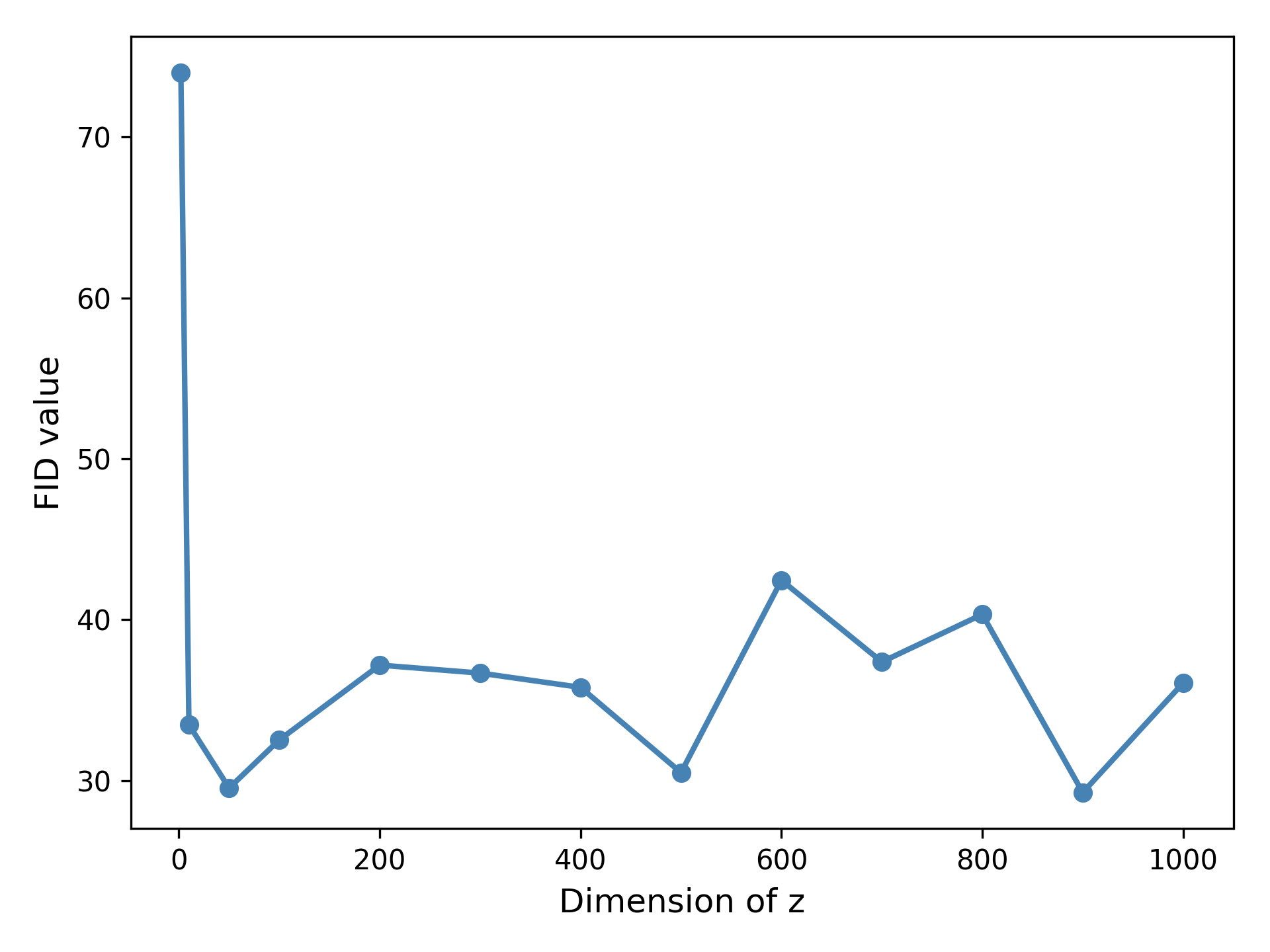}
        \caption{FID Plot}
        \label{fig:my_label}
    \end{subfigure} %
    \begin{subfigure}{.24\textwidth}
        \centering
        \includegraphics[width=.9\linewidth]{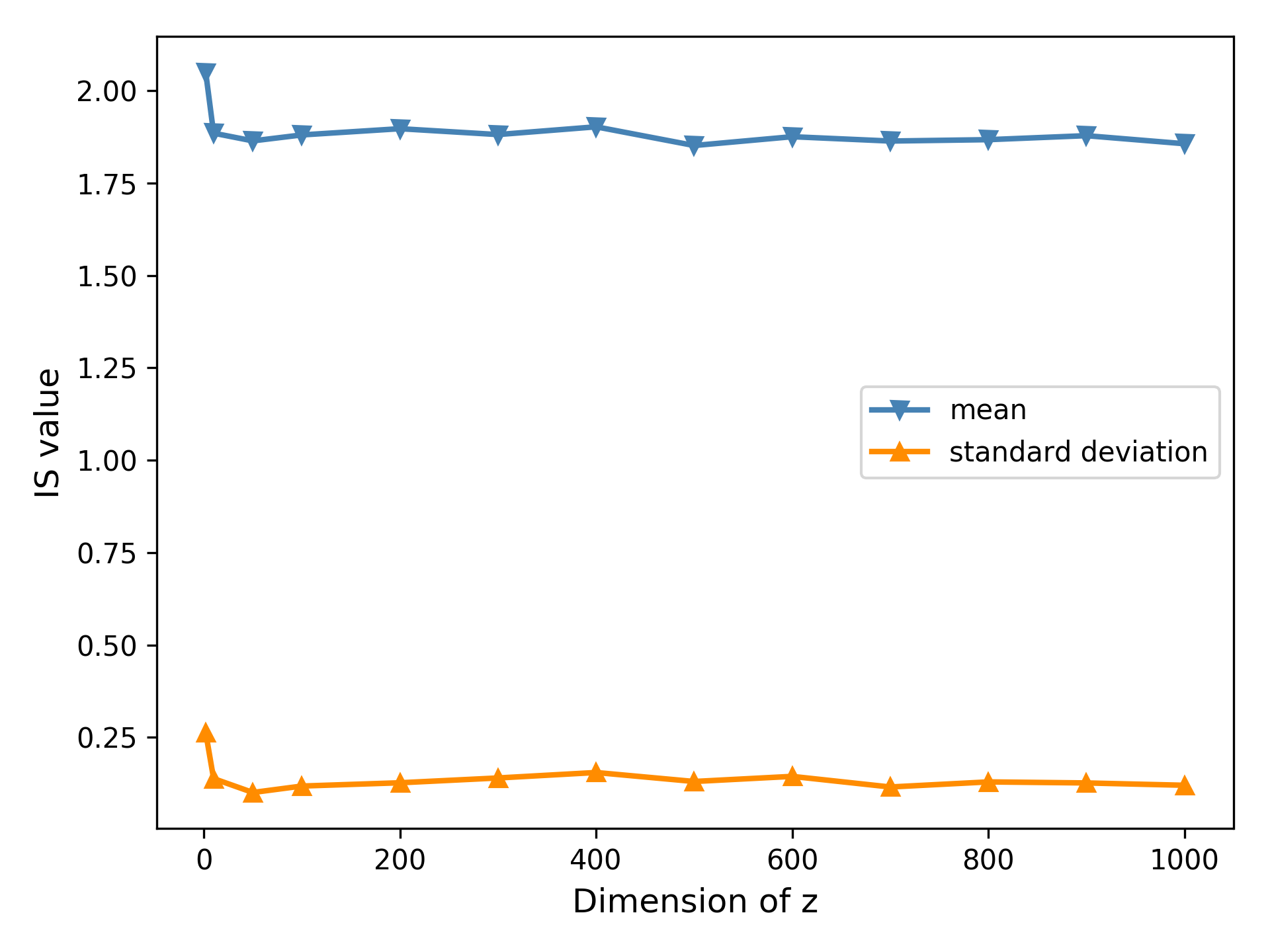}
        \caption{IS Plot}
        \label{fig:my_label}
    \end{subfigure}
\caption{Performance Measure Plots for 64x64 CelebA}
\label{fig:celeba4}
\end{figure}

In Figure \ref{fig:celeba4} (a) and (c), we compare the FID and IS scores for generating CelebA images for different dimensions of input noise for DCGAN and WGAN-GP. We fix the architecture of the generator and the discriminator and we find that the FID scores are very high for very low noise dimension but decreases considerably after a threshold and then do not change much for higher dimensions. We find the FID values for WGAN-GP are way better as also indicated by the figure. In Figure \ref{fig:celeba4} (b) and (d), we plot the IS values which are evaluated for batches of samples and the mean and the variance of the IS values across the batches is plotted for both DCGAN and WGAN-GP. The IS values do not seem to be indicative of the results as much as the FID values. WGAN-GP performs better at celebaA 64 than celebA 32 while the opposite is true for DCGAN.

\begin{figure}[!htb]
\centering
    \begin{subfigure}{.24\textwidth}
        \centering
        \includegraphics[width=.9\linewidth]{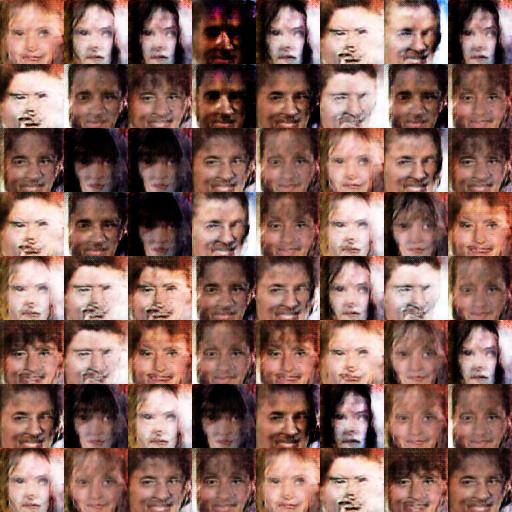}
        \caption{Dimension of noise = 2}
        \label{fig:my_label}
    \end{subfigure}  fo%
    \begin{subfigure}{.24\textwidth}
        \centering
        \includegraphics[width=.9\linewidth]{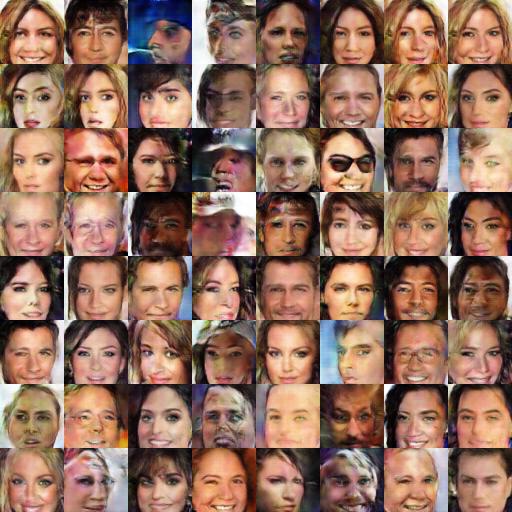}
        \caption{Dimension of noise = 10}
        \label{fig:my_label}
    \end{subfigure} %
    \begin{subfigure}{.24\textwidth}
        \centering
        \includegraphics[width=.9\linewidth]{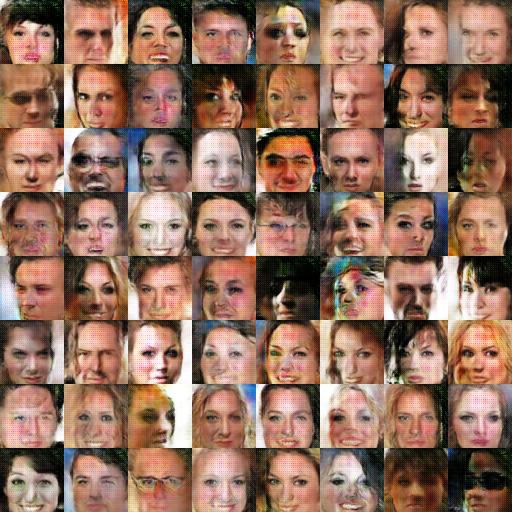}
        \caption{Dimension of noise = 100}
        \label{fig:my_label}
    \end{subfigure} %
    \begin{subfigure}{.24\textwidth}
        \centering
        \includegraphics[width=.9\linewidth]{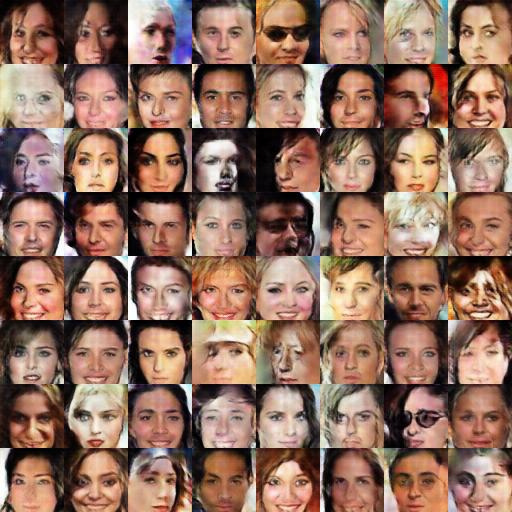}
        \caption{Dimension of noise = 500}
        \label{fig:my_label}
    \end{subfigure} %
\caption{Images Generated by DCGAN for 64x64 CelebA}
\label{fig:celeba5}
\end{figure}

\begin{figure}[!htb]
\centering
    \begin{subfigure}{.24\textwidth}
        \centering
        \includegraphics[width=.9\linewidth]{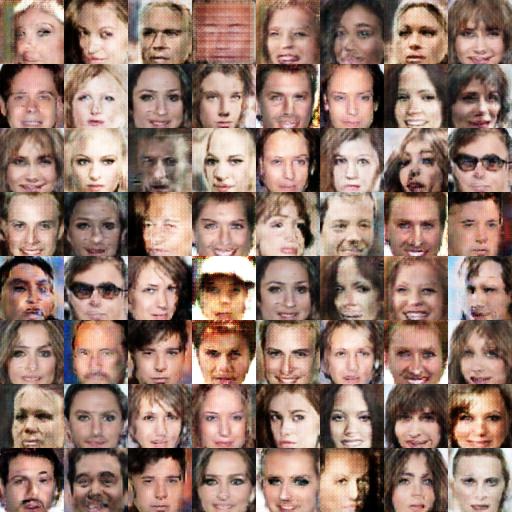}
        \caption{Dimension of noise = 2}
        \label{fig:my_label}
    \end{subfigure} %
    \begin{subfigure}{.24\textwidth}
        \centering
        \includegraphics[width=.9\linewidth]{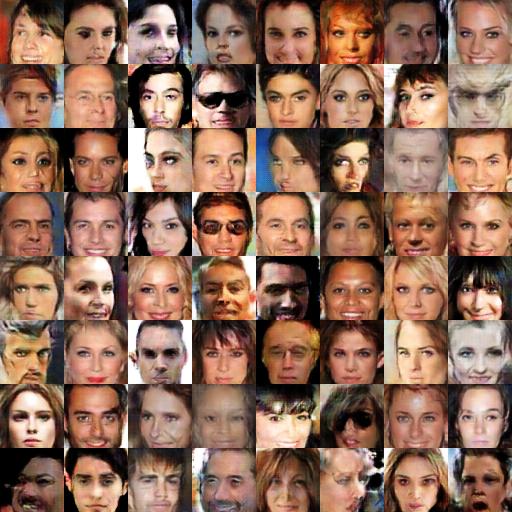}
        \caption{Dimension of noise = 10}
        \label{fig:my_label}
    \end{subfigure} %
    \begin{subfigure}{.24\textwidth}
        \centering
        \includegraphics[width=.9\linewidth]{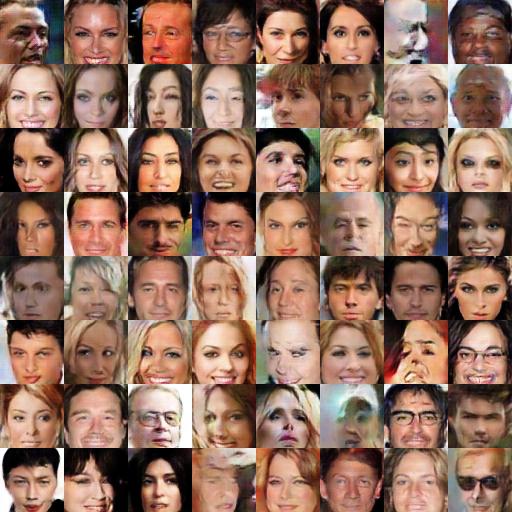}
        \caption{Dimension of noise = 100}
        \label{fig:my_label}
    \end{subfigure} %
    \begin{subfigure}{.24\textwidth}
        \centering
        \includegraphics[width=.9\linewidth]{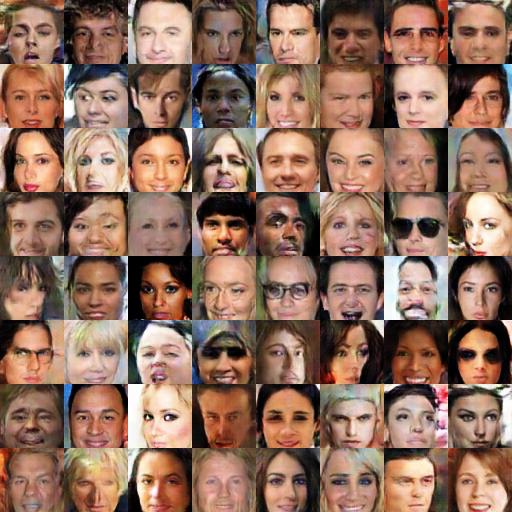}
        \caption{Dimension of noise = 500}
        \label{fig:my_label}
    \end{subfigure} %
\caption{Images Generated by WGAN-GP for 64x64 CelebA}
\label{fig:celeba6}
\end{figure}

We visually compare the results in Figures \ref{fig:celeba5}, \ref{fig:celeba6}. We find that results when noise dimension is 2 is bad compared to the other dimensions. WGAN-GP performs better compared to DCGAN. The clarity as well as variety of images generated by WGAN-GP is better compared to DCGAN unlike the case for images of size 32. 

\section{Discussion and Future Work}
\label{sec:disc}
In the various recent analysis on GANs, there has been hardly any focus on the input noise. The transformation of input noise to the generated data is a crucial part of the generation process. We think it calls for more attention and further analysis. 
From the results above, we can see that, there is a significant effect on the results when the input noise dimension is changed. It is also observed that the optimal noise dimension depends on the data-set and loss function used. 

Given that we intend to map the high dimensional data to a low dimensional distribution, we would like the dimension of noise to be as small as possible. Although very small values do not give good results, hence we find an optimal dimension after which the model does not perform better in case of CelebA and MNIST or performs worse in case of Gaussian data. 

FID value is not so indicative for analyzing the effect of change in input noise for the CelebA and MNIST data-set. We believe a more theoretical study and further analysis will help in training hand faster and better results than starting with a random size of z. We may also need to come up with performance measures which are more indicative of the quality of generated images.

\section{Conclusion}
\label{sec:conc}
We studied the effect of changing input noise for GANs quantitatively and qualitatively.  We conclude that the input noise dimension has a significant effect on the generation of images. To obtain useful and quality data generation, the input noise dimension needs to be set based on data-set and which GAN architecture used. We leave the theoretical analysis of the relation between the low dimensional distribution and the high dimensional data for the future work.



\bibliographystyle{acm}
\bibliography{noise}
\end{document}